\documentclass{article}
\usepackage{ijcai24}

\usepackage{times}
\usepackage{soul}
\usepackage{url}
\usepackage[hidelinks]{hyperref}
\usepackage[utf8]{inputenc}
\usepackage[small]{caption}
\usepackage{graphicx}
\usepackage{amsmath}
\usepackage{amsthm}
\usepackage{amsfonts}       
\usepackage{booktabs}
\usepackage{algorithm}
\usepackage{algorithmic}
\usepackage[switch]{lineno}
\usepackage{bm}             
\usepackage{xcolor}
\usepackage{pifont}
\usepackage{colortbl}
\usepackage{subcaption}

\definecolor{ourmethod}{gray}{0.93}
 
\newcommand{\T}{\mathbf{T}} 
\newcommand{\lam}{\bm{\lambda}} 

\newcommand{\cmark}{\textcolor{black}{\ding{51}}}
\newcommand{\xmark}{\textcolor{black}{\ding{55}}}  

\newtheorem{definition}{Definition}

\newtheorem{theorem}{Theorem}
\newtheorem{lemma}{Lemma}

\title{Learning Causally Disentangled Representations via the Principle of Independent Causal Mechanisms}

\author{
Aneesh Komanduri$^1$
\and
Yongkai Wu$^2$\and
Feng Chen$^{3}$\And
Xintao Wu$^1$ \\
\affiliations
$^1$University of Arkansas\\
$^2$Clemson University\\
$^3$University of Texas at Dallas
\emails
\{akomandu, xintaowu\}@uark.edu,
yongkaw@clemson.edu,
feng.chen@utdallas.edu
}

\begin{document}

\maketitle

\begin{abstract}
Learning disentangled causal representations is a challenging problem that has gained significant attention recently due to its implications for extracting meaningful information for downstream tasks. In this work, we define a new notion of causal disentanglement from the perspective of independent causal mechanisms. We propose ICM-VAE, a framework for learning causally disentangled representations supervised by causally related observed labels. We model causal mechanisms using nonlinear learnable flow-based diffeomorphic functions to map noise variables to latent causal variables. Further, to promote the disentanglement of causal factors, we propose a causal disentanglement prior learned from auxiliary labels and the latent causal structure. We theoretically show the identifiability of causal factors and mechanisms up to permutation and elementwise reparameterization. We empirically demonstrate that our framework induces highly disentangled causal factors, improves interventional robustness, and is compatible with counterfactual generation.
\end{abstract}

\section{Introduction}
Disentangled representation learning aims to learn meaningful and compact representations that capture semantic aspects of data by structurally disentangling the factors of variation \cite{bengio2014representation}. Such representations have been shown to offer useful properties such as better interpretability, robustness to distribution shifts, efficient out-of-distribution sampling, and fairness \cite{pmlr-v97-locatello19a}. However, disentangled representation learning typically assumes that the underlying factors are independent, which is unrealistic in practice. The factors generating the data can contain correlations or even causal relationships that are disregarded when factors are assumed to be independent. Further, a generative model learning from an independent prior assumes that all combinations of the latent factors are equally likely to appear in the training data. Thus, disentangling the factors would yield a sub-optimal likelihood since the assumed support could be well outside the support of the training data. 

\begin{figure}
\centering
\includegraphics[scale=0.5]{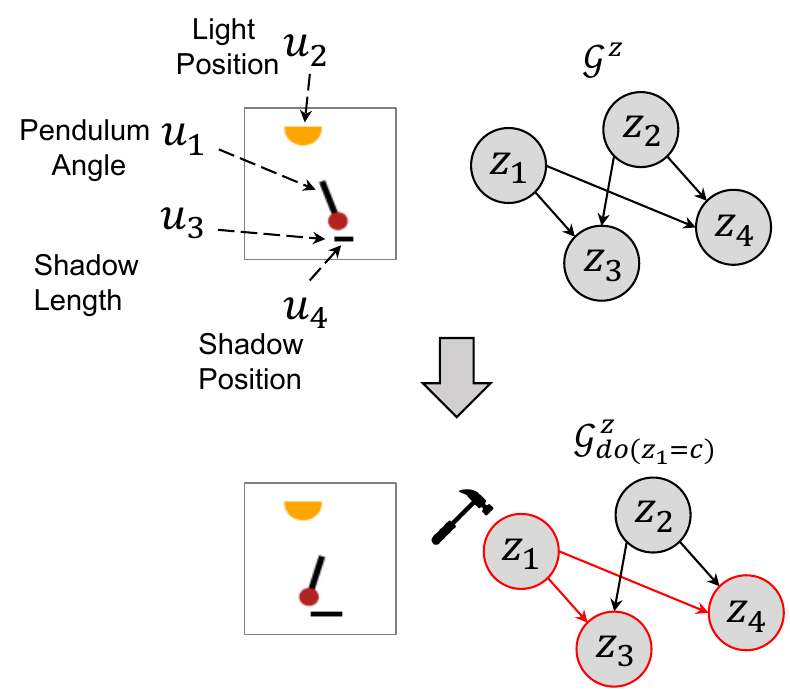}
\caption{We learn causal models representing images as latent causal variables $z$. The bottom shows the effect of intervening on the latent code corresponding to the pendulum's angle, propagating effects, and generating a counterfactual image.}
\label{fig:motivation_ex}
\end{figure}

Recently, there has been a growing interest in bridging causality \cite{Pearl09} and representation learning \cite{bengio2014representation}. The goal of causal representation learning (CRL) is to map unstructured low-level data to high-level abstract causal variables of interest \cite{scholkopf_toward_2021}. The key assumption is that high-dimensional observations are generated from a set of underlying low-dimensional \textit{causally related} factors of variation. Causal representations have been shown to be useful for tasks involving reasoning and planning. Causal representations also adhere to the principle of independent causal mechanisms (ICM) \cite{pmlr-v80-parascandolo18a}, which states that the mechanisms that generate each causal variable are independent such that a change in one mechanism does not affect another \cite{scholkopf_stat,10.5555/3202377}. Learning a generative model that captures the causal structure among latent factors can be crucial for reasoning about the world under manipulation. For example, a pendulum, light source, and shadow, as seen in Figure \ref{fig:motivation_ex}, may be causally related but are separate entities in the world that can be independently manipulated. Particularly, manipulating the pendulum's angle will affect the shadow's position and length. These hypothetical scenarios could be counterfactually generated from a causal model.

Ensuring that causal representations are \textit{disentangled} is useful for the causal controllable generation to generate counterfactual instances unseen during training. Such instances could be utilized as augmented data for robust learning in downstream tasks. The notion of disentanglement may be trivial when the factors are independent but becomes difficult to achieve when there are correlations or causal relationships among factors in the observed data. For highly correlated factors, it can be difficult to separate the factors of variation from their latent codes \cite{pmlr-v139-trauble21a}. Recently, it was shown that it is impossible to learn a disentangled representation in an unsupervised manner without some form of inductive bias \cite{pmlr-v97-locatello19a}. Recent work proved that models with an independent prior are unidentifiable \cite{shen_disentangled_2021}. Further, most existing disentanglement methods fail to disentangle factors when correlations exist in the data \cite{pmlr-v139-trauble21a}. However, results from large-scale empirical studies \cite{locatello_weakly-supervised_2020} have indicated that supervision in the form of auxiliary labels or contrastive data can effectively disentangle correlated or causal factors.

\paragraph{Related Work.} Our work builds upon the ideas presented in iVAE \cite{pmlr-v108-khemakhem20a} and causal variants \cite{DBLP:conf/cvpr/YangLCSHW21,komanduri-scm-vae} and extends them to consider a principled view of causal disentanglement in the label supervised setting. DIVA \cite{diva} and CCVAE \cite{joy2021capturing} are special case implementations of the iVAE framework. Yang et al. \shortcite{DBLP:conf/cvpr/YangLCSHW21} proposed CausalVAE, which uses a causal masking layer and is limited to linear SCMs. Komanduri et al. \shortcite{komanduri-scm-vae} extended this to a nonlinear setting and proposed a causal prior. Both works proposed simplistic models to learn causal mechanisms under the strictly additive noise assumption and do not, from an empirical or theoretical perspective, focus on disentanglement. Shen et al. \shortcite{shen_disentangled_2021} proposed learning causal representations supervised by a GAN loss. There has also been work focusing on learning causal representations from paired counterfactual data \cite{brehmer2022weakly}, temporal data \cite{iCITRIS}, in self-supervised learning \cite{self_supervised}, and when interventional data is available \cite{ahuja_interventional}. Unlike many previous works in label-supervised VAE-based CRL, we consider general nonlinear SCMs instead of restricting to additive noise models, propose a causal prior to causally factorize the latent space, and achieve disentanglement of causal mechanisms, as summarized in Table \ref{properties}.

\paragraph{Contributions.} (1) We propose ICM-VAE, a framework for causal representation learning under label supervision, where causal variables are derived from nonlinear flow-based diffeomorphic causal mechanisms. (2) Based on the ICM principle, we propose the notion of causal disentanglement for causal models from the perspective of mechanisms and design a causal disentanglement prior to causally factorize the learned distribution over causal variables. (3) Using the structure from our causal disentanglement prior, we theoretically show identifiability of the learned causal factors up to permutation and elementwise reparameterization. (4) We experimentally validate our method and show that our model can almost perfectly disentangle the causal factors, improve interventional robustness, and generate consistent counterfactuals.

\begin{table}[t]
\centering
\begin{minipage}[b]{\linewidth}
\centering
\resizebox{\textwidth}{!}{
\begin{tabular}{lcccc}
\toprule
{\bf Framework} &
  \begin{tabular}[c]{@{}c@{}}\bf General SCM \\ \bf Compatible \end{tabular} &
  \begin{tabular}[c]{@{}c@{}}\bf Causal \\ \bf Prior \end{tabular} &
  \begin{tabular}[c]{@{}c@{}}\bf Causal Mechanism  \\ \bf Disentanglement\end{tabular}
  \\ \hline
{iVAE} & \xmark & \xmark & \xmark  \\
{CausalVAE} & \xmark & \xmark & \xmark \\
{SCM-VAE}  & \xmark & \cmark & \xmark \\
\rowcolor{ourmethod} {ICM-VAE (Ours)}  & \cmark & \cmark & \cmark \\
\bottomrule
\end{tabular}
}
\end{minipage}
\caption{Causal and acausal framework compatibilities in label-supervised setting}
 \label{properties}
\end{table}

\section{Preliminaries}

Let $\mathcal{X} \subset \mathbb{R}^d$ denote the support of the observed data $x$ generated from a set of causally related ground-truth factors. The observations are assumed to be explained by some latent causal factors of variation $z$ with domain $\mathcal{Z} \subset \mathbb{R}^n$, where $n \ll d$. 
$z$ represents a set of causal factors while $z_i$ represents a single causal factor.
We assume $x$ can be decomposed as $x = g(z) + \xi$ where $\xi \sim \mathcal{N}(0, \sigma^2 I)$ are mutually independent Gaussian noise terms for reconstruction. Let $g: \mathcal{Z} \to \mathcal{X}$ be the decoder (or mixing function). Each factor $z_i$ contains semantically meaningful information about the observation. In the traditional VAE \cite{kingma2013}, we assume the observed data is generated by the latent generative model with the structure  $p_{\theta}(x,z) = p_{\theta}(x|z) p_{\theta}(z)$ where $\theta$ are the true but unknown parameters. We aim to learn the joint distribution $p(x, z)$ to estimate the marginal density and a posterior $p(z|x)$ to describe the underlying factors of variation given a prior $p(z)$ over the latent variables. 

\subsection{Identifiability}
The goal of learning a useful representation that recovers the true underlying data-generating factors is closely tied to the problem of blind source separation (BSS) and independent component analysis (ICA) \cite{hyvarinen-auxiliary,COMON1994287,HYVARINEN1999429}. Provably showing that a learning algorithm achieves this goal up to tolerable ambiguities under certain conditions is formalized as the \textit{identifiability} of a model. In this section, we use the notion of $\sim$-equivalence \cite{pmlr-v108-khemakhem20a} to formulate the notion of identifiability.

\begin{definition}[$\sim$-identifiability]
\label{ident}
Let $\sim$ be an equivalence relation on $\theta$. The generative model is $\sim$-identifiable if 
\begin{equation}
    p_{\theta}(x) = p_{\hat{\theta}}(x) \implies \theta \sim \hat{\theta}
\end{equation}
\end{definition}

If two different choices of model parameter $\theta$ and $\hat{\theta}$ lead to the same marginal density $p_{\theta}(x)$, then they must be equal and this implies that $p_{\theta}(x,z)= p_{\hat{\theta}}(x,z)$, $p_{\theta}(z)= p_{\hat{\theta}}(z)$, and $p_{\theta}(z|x)= p_{\hat{\theta}}(z|x)$. However, recent work showed that it is impossible to achieve marginal density equivalence $p_{\theta}(x) = p_{\hat{\theta}}(x)$ with an unconditional prior $p_{\theta}(z)$ \cite{pmlr-v108-khemakhem20a}. Since the VAE is unidentifiable without some form of additional restriction on the function class of the mixing function or auxiliary information, the identifiable VAE (iVAE) was proposed to utilize auxiliary information in the form of a conditionally factorial prior for identifiability guarantees \cite{pmlr-v108-khemakhem20a}.
In iVAE, each factor $z_i$ is assumed to have a univariate exponential family distribution (due to their universal approximation capabilities) given the conditioning variable $u$, where a function $\lam$ determines the natural parameters of the distribution. The general PDF of the conditional distribution is defined as follows:
\begin{equation}
    \resizebox{.91\linewidth}{!}{$
            \displaystyle
        p_{T, \lambda}(z | u) = \prod_i h_i(z_i) \exp\Bigg[\sum_{j=1}^k T_{i, j}(z_i) \lambda_{i, j}(u) - \psi_i(u)\Bigg]
            $}
    \label{eq:general_conditional_prior}
\end{equation}

where $h_i(z_i)$ is the base measure, $\T_i: \mathcal{Z} \to \mathbb{R}^k$ and $\T_i = (T_{i, 1}, \dots, T_{i, k})$ are the sufficient statistics, $\lam_i(u) = (\lambda_{i, 1}(u), \dots, \lambda_{i, k}(u))$ are the corresponding natural parameters, $k$ is the dimension of each sufficient statistic, and the remaining term $\psi_i(u)$ acts as a normalizing constant. A prior conditioned on auxiliary information $u$ can guarantee that the joint densities $p_{\theta}(x, z) = p_{\hat{\theta}}(x, z)$ are equivalent up to some equivalence class. The following two definitions describe the conditions necessary to achieve the identifiability of a learned model up to linear transformation and block permutation indeterminacies, respectively.

\begin{definition}[\cite{pmlr-v108-khemakhem20a}]
\label{lin_eq}
Let $\sim$ be an equivalence relation on $\theta$, $\mathcal{X} = g(\mathcal{Z})$, and $\mathcal{\hat{X}} = \hat{g}(\mathcal{Z})$. We say that $\theta$ and $\hat{\theta}$ are \textbf{linearly-equivalent} if and only if there exists an invertible matrix $A\in \mathbb{R}^{nk \times nk}$ and vectors $b, c\in \mathbb{R}^{nk}$ such that $\forall x\in \mathcal{X}$:
\begin{enumerate}
    \item $\T(g^{-1}(x)) = A\hat{\T}(\hat{g}^{-1}(x)) + b, \forall x\in \mathcal{X}$
    \item $A^T\lam(u) + c = \hat{\lam}(u)$
\end{enumerate}
We denote this equivalence as $\theta \sim_A \hat{\theta}$.
\end{definition}

\begin{definition}[\cite{pmlr-v108-khemakhem20a}]
\label{perm_eq}
We say $\theta$ and $\hat{\theta}$ are \textbf{permutation-equivalent}, denoted $\theta \sim_P \hat{\theta}$, if and only if $P$ is permutation matrix that has block-permutation structure respecting $\T$. That is, there exist $n$ invertible $k\times k$ matrices $A_1, \dots, A_n$ and an $n$-permutation $\pi$ such that for all $z\in \mathbb{R}^{nk}$, $P \hat{z} = [z_{\pi(1)}A_1^T, z_{\pi(2)}A_2^T, \dots, z_{\pi(n)}A_n^T]^T$. 
\end{definition}

Linear equivalence indicates the true representation is a linear transformation of the learned representation and only guarantees the learned representation captures the true representation. In general, linear-equivalent identifiability does not guarantee that the factors of variation are disentangled since the linear transformation can mix up the variables (i.e. one component of $g^{-1}$ corresponds to multiple components of $\hat{g}^{-1}$). Permutation equivalence implies that the $i$-th factor $z_i$ of one representation corresponds to a unique factor in another representation, given the permutation $\pi$. To truly disentangle factors of variation, we must ensure that each coordinate of the learned representation is equal to the scaled and shifted coordinate of the ground truth up to some permutation. To this end, we define the notion of disentanglement as permutation equivalence \cite{lachapelle2022disentanglement} as follows.

\begin{definition}[Permutation Disentanglement] 
\label{disentanglement}
Given some ground-truth model, a learned model $\hat{\theta}$ is said to be \textbf{disentangled} if $\theta$ and $\hat{\theta}$ are permutation-equivalent.
\end{definition}

\subsection{Structural Causal Model}
Henceforth, we assume $z$ is described by a structural causal model (SCM) \cite{Pearl09}, which is formally defined by a tuple $\mathcal{M} = \langle \mathcal{Z}, \mathcal{E}, F\rangle$, where $\mathcal{Z}$ is the domain of the set of $n$ endogenous causal variables $z=\{z_1, \dots, z_n\}$, $\mathcal{E}$ is the domain of the set of $n$ exogenous noise variables $\epsilon=\{\epsilon_1, \dots, \epsilon_n\}$, which is learned as an intermediate latent variable, and $F = \{f_1, \dots, f_n\}$ is a collection of $n$ independent causal mechanisms of the form
\begin{equation}
    z_i = f_i(\epsilon_i, z_{\textbf{pa}_i})
\label{eq:prelim_scm}
\end{equation}
where $\forall i$, $f_i: \mathcal{E}_i \times \prod_{j\in \textbf{pa}_i} \mathcal{Z}_j \to \mathcal{Z}_i$ are \textbf{causal mechanisms} that determine each causal variable as a function of the parents and noise, $z_{\textbf{pa}_i}$ are the parents of causal variable $z_i$; and a probability measure $p_{\mathcal{E}}(\epsilon) = p_{\mathcal{E}_1}(\epsilon_1)p_{\mathcal{E}_2}(\epsilon_2)\dots p_{\mathcal{E}_n}(\epsilon_n)$, which admits a product distribution. For the purposes of this work, we assume a causally sufficient setting (no hidden confounding), no SCM misspecification, and faithfulness is satisfied. We depict the causal structure of $z$ by a directed acyclic graph (DAG) $\mathcal{G}^z$ with adjacency matrix $G^z \in \{0, 1\}^{n\times n}$.

\section{Causal Mechanism Equivalence}
\label{sec:theory}

Although the existing notions of disentanglement may be suitable for independent factors of variation \cite{pmlr-v108-khemakhem20a}, they fail to capture important information in a causal model where the factors are causally related. As formulated in Def. \ref{lin_eq} and Def. \ref{perm_eq}, linear equivalence or permutation equivalence cannot capture the causal mechanisms accurately or distinguish the mechanisms afflicted to factors. For a counterexample to the definitions, refer to Appendix \ref{app:counterex}. The framework of iVAE captures identifiability in the sense that the joint distributions of the latent variables of two different models are equivalent. However, for a causally factorized model, we have that $p_{\theta}(z) = p_{\hat{\theta}}(z)$ does not imply $p_{\theta}(z_i | z_{\textbf{pa}_i}) = p_{\hat{\theta}}(z_i | z_{\textbf{pa}_i})$. That is, the ground-truth causal factors and the learned causal factors should entail the same causal conditional mechanisms, where the minimal conditioning set is the set of causal parents. 

Based on the intuition that causal models are described by mechanisms, we define a new notion of disentanglement that takes into account conditional distributions of causal variables under the Markov factorization. The new causal conditional equivalence preserves information about the independent causal mechanisms (ICM), which is a unique formulation for a causal model and important for performing correct interventions. The following two definitions describe the conditions necessary to satisfy causal mechanism equivalence. 

\begin{definition}[Causal Mechanism Permutation Equivalence]
\label{causal_perm}
Let $\sim$ be an equivalence relation between $\hat\theta$ and $\theta$, $\mathcal{X} = g(\mathcal{{Z}})$, and $\mathcal{\hat{X}} = \hat{g}(\mathcal{{Z}})$.
If the factors $z$ are causally related, we say that $\theta$ is causal mechanism permutation equivalent to $\hat{\theta}$ iff:
    \begin{enumerate}    
    \item There exists a permutation matrix $P$ such that $I = P\cdot J$ where $I$ and $J$ are indices of $z$ and $\hat{z}$, respectively.
    \item Given an equivalence pair $(i, j)$, i.e., $P_{ij} \neq 0$, from this permutation matrix, one has $\T_i(z_i | z_{\textbf{pa}_i}) = D_{ij} \hat{\T}_j(z_j | z_{\textbf{pa}_j}), \forall z_i \in  \mathcal{Z}_i, \forall \hat{z}_j \in  \mathcal{Z}_j$, where $D_{ij}$ is a scaling coefficient.
    \item For all $i, j\in \{1, \dots, n\}$, we have the mechanism equivalence $\lam_i(z_{\textbf{pa}_i}, u) = D_{ij}\hat{\lam}_j(z_{\textbf{pa}_j}, u)$, where $D$ is a diagonal scaling matrix.
\end{enumerate}
\end{definition}

\begin{definition}[Causal Disentanglement] 
\label{cdisentanglement}
Given some ground-truth model $\theta$, a learned model $\hat{\theta}$ is said to be \textbf{causally disentangled} if $\theta$ and $\hat{\theta}$ are causal mechanism permutation-equivalent.
\end{definition}

\begin{figure*}
    \centering
    \includegraphics[width=0.97\textwidth]{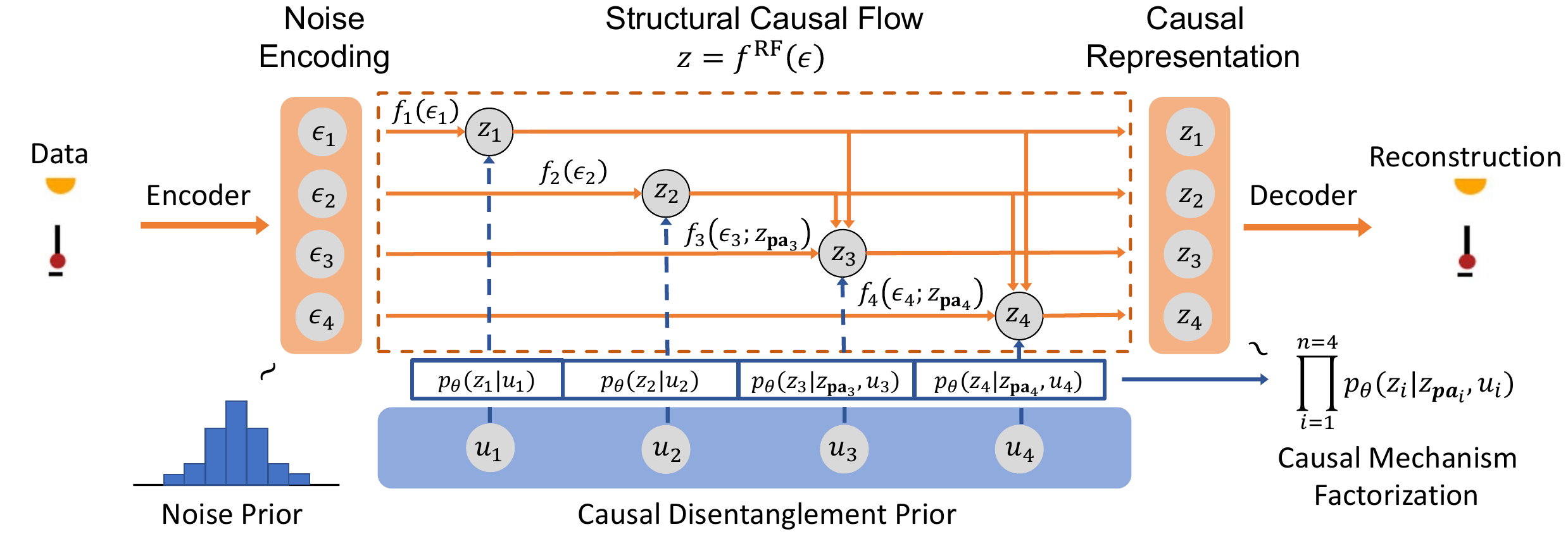}
    \caption{Architecture of ICM-VAE Framework, which contains two main components: (i) Structural Causal Flow (SCF), and (ii) Causal Disentanglement Prior. The blue color represents prior components and the orange represents the learning process.}
    \label{fig:framework}
\end{figure*}

\section{Proposed Framework}
We design a framework to achieve \textit{causal} disentanglement. We propose ICM-VAE, a VAE-based framework based on the independent causal mechanisms (ICM) principle that achieves disentanglement of causal mechanisms. Figure \ref{fig:framework} shows the overall architecture of our proposed framework.

\subsection{Structural Causal Flow}
Rather than assuming the limiting linear causal graphical model (CGM), as done in CausalVAE \cite{DBLP:conf/cvpr/YangLCSHW21}, we consider causal mechanisms to be complex nonlinear functions. Diverging from the strictly additive noise model assumption, we propose to parameterize causal mechanisms with a more general \textit{diffeomorphic}\footnote{A diffeomorphism is a differentiable bijection with a differentiable inverse.} function. Flow-based models \cite{norm_flow} are often quite expressive in low-dimensional settings, which makes them desirable for learning complex distributions due to efficient and exact evaluation of densities. We parameterize the causal mechanisms with a conditional flow, which we refer to as the latent structural causal flow (SCF), that learns to map the independent noise distribution to a distribution over causal variables. This module is inspired by the causal autoregressive flow \cite{pmlr-v130-khemakhem21a}. This type of model is more realistic and general to better capture the complex distribution over the latent causal variables compared to simple linear mappings and leads to counterfactual identifiability \cite{nasresfahany2023counterfactual}. The SCF, denoted as $f^{\text{RF}}$, is the reduced form (RF) of a nonlinear SCM function that conceptually maps noise variables $\epsilon$ to causal variables $z$ as follows
\begin{equation}
    z = f^{\text{RF}}(\epsilon)
    \label{eq:reduced_form_scm}
\end{equation}

where $f^{\text{RF}}: \mathcal{E} \to \mathcal{Z}$ is derived from the recursive substitution of causal mechanisms $f_i$ in topological order of the causal graph as follows
\begin{equation}
z_i = f_i(\epsilon_i; z_{\textbf{pa}_i}), \qquad \forall i\in \{1, \dots, n\}    
\label{eq:recursive_scm}
\end{equation}
 realized as a function of the noise term and parent variables. The noise encoding $\epsilon_i$ is exactly the SCM noise variable corresponding to the causal variable $z_i$. 
 
Similar to several prior works \cite{shen_disentangled_2021,liang2023causal}, we assume that the latent causal graph is known in the form of a binary adjacency matrix to focus on formulating the problem of causal disentanglement. To implement a diffeomorphic function $f^{\text{RF}}$, we use flow-based models to parameterize the causal mechanisms. Specifically, this flow is implemented as an affine-form autoregressive flow, where we derive each causal variable one at a time in topological order such that each variable is dependent only on a subset of previously derived variables (i.e. parents). Thus, the change of variables can be computed quite easily for exact and efficient likelihood estimation. In general, one can parameterize causal mechanisms using any nonlinear diffeomorphic function, as long as it takes into account the topological ordering. Let's take the pendulum example in Figure \ref{fig:motivation_ex} to illustrate. The causal structure is $z_1 \to z_3, z_4$ and $z_2 \to z_3, z_4$. 
Then, the SCF would be $f^{\text{RF}}: (\epsilon_1, \epsilon_2, \epsilon_3, \epsilon_4) \mapsto (z_1 = f_1(\epsilon_1), z_2 = f_2(\epsilon_2), z_3 = f_3(\epsilon_3, z_1, z_2), z_4 = f_4(\epsilon_4, z_1, z_2))$, where $z_i = f^{\text{RF}}_i(\epsilon_i; \epsilon_{\textbf{pa}_i}) = f_i(\epsilon_i; z_{\textbf{pa}_i})$ and each $f_i$ is a diffeomorphic transformation of the form
\begin{equation}
    z_i = f_i(\epsilon_i; z_{\textbf{pa}_i}) = \exp(a_i) \cdot \epsilon_i + b_i
    \label{eq:scf_affine_form}
\end{equation}

where $a_i = r_1(z_{\textbf{pa}_i})$ and $b_i = r_2(z_{\textbf{pa}_i})$ are the slope and offset parameters of the affine transformation, respectively, learned via neural networks $r_1$ and $r_2$ that capture information about the causal parents. Since the Jacobian of the function will be triangular by construction and the slope parameter is learned for each variable, the slope is equivalent to the diagonal elements of the Jacobian matrix as follows
\begin{equation}
    \log \prod_{i} \bigg|\frac{\partial \epsilon_i}{\partial z_i}\bigg| = \sum_{i} \log \bigg|\frac{\partial f^{\text{RF}}_i(\epsilon_i; \epsilon_{\textbf{pa}_i})}{\partial \epsilon_i}\bigg|^{-1} = \sum_{i} a_i
    \label{eq:scf_logdet}
\end{equation}

where $\epsilon_{\textbf{pa}_i}$ denotes the noise terms associated with the parents of causal variable $z_i$. The structural causal flow can easily be generalized to multivariate scenarios by masking groups of latent codes corresponding to each causal variable. 

\subsection{Generative Model}
To achieve an identifiable model, we leverage auxiliary information as a weak supervision signal \cite{pmlr-v108-khemakhem20a}. Let $u\in \mathbb{R}^n$ be the auxiliary observed labels corresponding to the causally related ground-truth factors with support $\mathcal{U}\subset \mathbb{R}^n$. We assume that the decoder $g$ is diffeomorphic onto its image. Several prior works \cite{locatello_weakly-supervised_2020,pmlr-v108-khemakhem20a,lachapelle2022disentanglement} assume that the nonlinear mixing function mapping $\mathcal{Z}$ to $\mathcal{X}$ is a diffeomorphism. 
Consider the pendulum system from Figure \ref{fig:motivation_ex} consisting of a light source, a pendulum, and a shadow. Given only the image, it is completely certain that we can identify where each object appears in the image. So, we find it reasonable to assume a diffeomorphic mixing function $g$ for our exploration. Let $\theta = (g, \T, \lam, G^z)$ be the parameters of the conditional generative model defined as follows
\begin{equation}
    p_{\theta}(x, \epsilon, z | u) = p_{\theta}(x | \epsilon, z)p_{\theta}(\epsilon, z | u)
    \label{generative_model}
\end{equation}
where 
\begin{equation}
    p_{\theta}(x | \epsilon, z) = p_{\theta}(x | z) = p_{\xi}(x - g(z))
    \label{likelihood}
\end{equation}
 If we assume that the distribution over the noise $\xi$ is Gaussian with infinitesimal variance, we can model non-noisy observations as a special case of Eq. (\ref{likelihood}). The prior distribution in the generative model is given by 
\begin{equation}
    p_{\theta}(\epsilon, z | u) = p(\epsilon)p_{\theta}(z | u)
    \label{joint_prior}
\end{equation} 
where we choose $p(\epsilon)$ as a standard Gaussian base distribution and $p(z | u)$ is assumed to be conditionally factorial. However, the conditional prior in Eq. (\ref{eq:general_conditional_prior}) cannot properly capture causal mechanisms for causally related factors. We next define a causally factorized prior suitable to achieve causal disentanglement.

\subsection{Causal Disentanglement Prior}
\label{disent_prior} We aim to use a structured prior and perform conditioning in the latent space, similar to previous work on nonlinear ICA \cite{pmlr-v108-khemakhem20a}, to enforce $z$ to be a disentangled causal representation. However, for a model incorporating causal structure, the form of the conditional prior in Eq. (\ref{eq:general_conditional_prior}) needs to be modified and generalized to \textit{causally} factorized distributions. To enforce the disentanglement of $z$, we parameterize the prior distribution to learn a mapping from $u$ to $z$. That is, since the goal of causal disentanglement is to map each latent/mechanism to exactly one corresponding ground-truth factor/mechanism, we can explicitly incorporate this into the prior. Using $u$ as our observational labels, we parameterize the factorized causal conditionals with a conditional flow between $u$ and $z$ to establish a bijective relationship. The goal is for the distribution over the causal variables to tend towards the learned prior. The prior over $z$ is defined as follows
\begin{equation}
\begin{split}
    p_{\theta}(z | u) &= \prod_{i=1}^n p_{\theta}(z_i | z_{\textbf{pa}_i}, u_i) = \prod_{i=1}^n p(u_i) \bigg|\frac{\partial \lam_i(u_i;z_{\textbf{pa}_i})}{\partial u_i}\bigg|^{-1}
    \label{eq:disentanglement_prior}
\end{split}
\end{equation}

\begin{align}
    p_{\theta}(z_i | z_{\textbf{pa}_i}, u_i) = h_i(z_i)\exp(\T_i(z_i | z_{\textbf{pa}_i})\lam_i(G_i^z &\odot z, u_i) \nonumber \\- \psi_i(z, u))
    \label{eq:disentanglement_prior_unit}
\end{align}

where $\lam_i(G_i^z \odot z, u_i)$ is the estimated parameter vector of the prior obtained via mechanism $\lam_i$, $G^z_i$ is the $i$th column of the adjacency matrix of the causal graph of $z$, $h_i(z)$ is the base measure, and $\T_i(z) = (z, z^2)$ is the sufficient statistic. The prior induces a causal factorization of $z$ with causal conditionals $p_{\theta}(z_i | z_{\textbf{pa}_i}, u_i)$, where $u_i$ is introduced as a weak supervision signal for identifiability. Eq. (\ref{eq:disentanglement_prior}) is reminiscent of temporal priors that define a distribution over a latent variable conditioned on the variable at a previous time step \cite{iCITRIS}. In our case, we view the causal factors as derived autoregressively. With a slight abuse of notation, we define $\lam(z, u)$ to be the concatenation of all $\lam_i(G_i^z \odot z, u_i)$. The function $\lam(z, u)$ outputs the natural parameter vector for the causally factorized distribution. We further require $\lam: \mathcal{Z} \times \mathcal{U} \to \mathcal{Z}$ to be a bijective map between $u$ and learned representation $z$ to encourage disentanglement of the causal mechanisms. In practice, we choose $p(u)$ from a location-scale family such as Gaussian. The learned mechanism $\lam_i$ is defined as the following diffeomorphic map:
\begin{equation}
    \lam_i(u_i; z_{\textbf{pa}_i}) = \exp(c_i) \cdot u_i + d_i
    \label{eq:dfp_affine_form}
\end{equation}
 where $c_i = s_1(z_{\textbf{pa}_i})$ and $d_i = s_2(z_{\textbf{pa}_i})$ are the slope and offset parameters of the flow, respectively, learned via neural networks. To obtain a causally factorized conditional prior over $z$, we map the base distribution $p(u)$, which is known beforehand, to a distribution over $z$.

\subsection{Learning Objective}
Putting all the components together, ICM-VAE consists of a stochastic encoder $q_{\phi}(\epsilon, z | x, u)$, a decoder $p_{\theta}(x | \epsilon, z)$, and diffeomorphic causal transformations $f_i(\cdot; \epsilon)$. All components are learnable and implemented as neural networks. Formally, we aim to optimize the following variational lower bound: 

\begin{align}
    \log p_{\theta}(x, u) \geq \mathbb{E}_{\epsilon, z \sim q_{\phi}(\epsilon, z | x, u)} \Big[\log p_{\theta}(x | \epsilon) &+ \log p_{\theta}(x| z) \nonumber \\ -  \beta \{\log q_{\phi}(\epsilon | x, u) +\log q_{\phi}(z | x, u) \nonumber \\- \log p(\epsilon) - \log p_{\theta}(z | u)\} \Big]
    \label{eq:dscmvae_elbo}
\end{align}

where $\beta$ is the latent bottleneck parameter. We train the model by minimizing the negative of the ELBO loss and learn to map low-level pixel data to noise variables and map the noise variable distribution to a distribution over the causal variables. For a detailed derivation of the ELBO, refer to Appendix \ref{app:elbo_der}. Causal structure learning could be heuristically incorporated jointly with the learning objective by enforcing acyclicity and sparsity of the causal graph. However, most causal discovery methods, such as GraN-DAG \cite{Lachapelle2020Gradient-Based}, require restricting parametric assumptions on the causal model (i.e., additive noise), to be practically applied. For an extended discussion on incorporating causal discovery and challenges, refer to Appendix \ref{app:discovery_discussion}.

\begin{figure*}
    \centering
    \includegraphics[width=\textwidth]{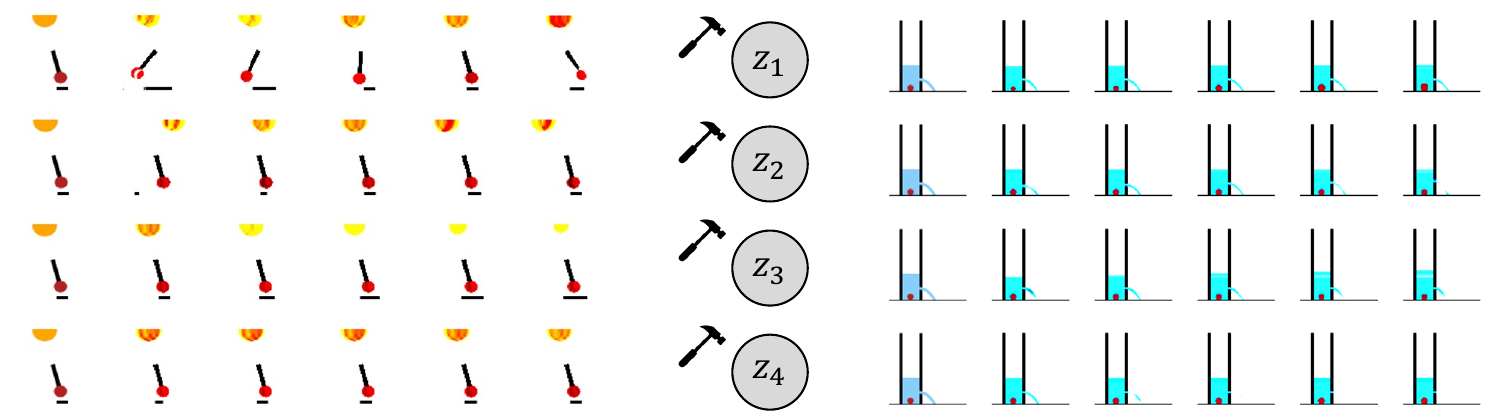}
    \caption{ Pendulum (left) and Flow (right) counterfactual images after intervening on causal factors, individually, and propagating effects.}
    \label{fig:pendflow_ctrfact}
\end{figure*}

\section{Identifiability Analysis}
We design our framework to satisfy the conditions necessary to achieve causal mechanism equivalence and causally disentangle the factors of variation. The causally factorized prior in Eq. (\ref{eq:disentanglement_prior}) induces disentanglement of causal mechanisms. Theorem \ref{thm1} extends the identifiability theorem from iVAE \cite{pmlr-v108-khemakhem20a} to show causal mechanism equivalence identifiability when we have a causal model. We note that causal mechanism disentanglement implies the disentanglement of causal factors.

\begin{theorem}[\normalfont{Identifiability of ICM-VAE}] \label{thm1} Suppose that we observe data sampled from a generative model defined according to (\ref{generative_model})-(\ref{eq:disentanglement_prior_unit}) with two sets of model parameters $\theta = (g, \T, \lam, G^z)$ and $\hat{\theta} = (\hat{g}, \hat{\T}, \hat{\lam}, \hat{G}^z)$. Suppose the following assumptions hold

\begin{enumerate}
    \item The set $\{x\in \mathcal{X} | \phi_{\xi}(x) = 0\}$ has measure zero, where $\phi_{\xi}$ is the characteristic function of the density $p_{\xi}$ defined in Eq. (\ref{likelihood}).
    \item The decoder $g$ is diffeomorphic onto its image.
    \item The sufficient statistics $\T_{i}$ are diffeomorphic.
    \item {\textbf{[Sufficient Variability]}} There exists $nk + 1$ distinct points $u_0, \dots, u_{nk}$ such that the matrix
    \begin{equation}
    \begin{split}
        L = (\lam(z_{\textbf{pa}_{(1)}}, u_{(1)}) - \lam(z_{(0)}, u_{(0)}), \dots,\\ \lam(z_{\textbf{pa}_{(nk)}}, u_{(nk)}) - \lam(z_{(0)}, u_{(0)}))
    \end{split}
    \end{equation}
    of size $nk \times nk$ is invertible, the ground-truth function $\lam$ is affected sufficiently strongly by each individual label $u_i$ and previously derived variables $z_{\textbf{pa}_i}$, and $\forall i$, $\lam_i(z_{\textbf{pa}_i}, u_i) \neq 0$.
\end{enumerate}
Then $\theta$ and $\hat{\theta}$ are causal mechanism permutation-equivalent, and the model $\hat{\theta}$ is causally disentangled. 
\end{theorem}

\begin{figure*}
    \centering
    \includegraphics[width=\textwidth]{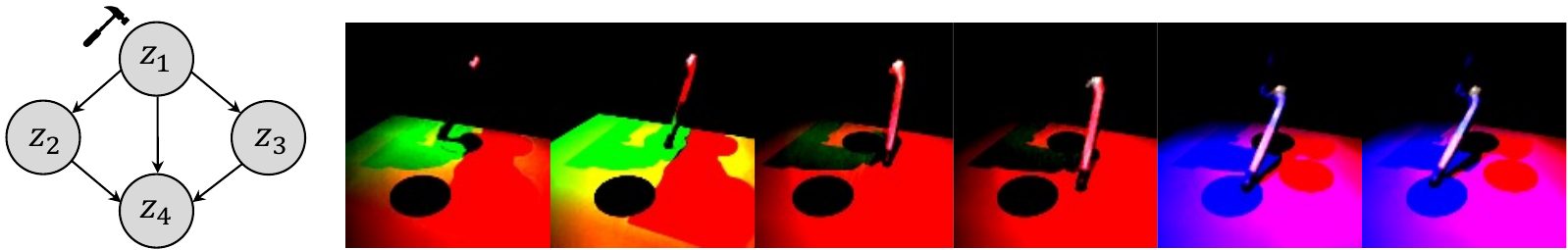}
    \caption{CausalCircuit counterfactual images by intervening on robot arm to turn on a colored light, which can causally affect other lights.}
    \label{fig:cc_ctrfact}
\end{figure*}
For the proof of Theorem 1, refer to Appendix \ref{app:proof}.

\section{Experimental Evaluation}
In this section, we empirically evaluate the effectiveness of ICM-VAE. We consider a setting where the causal variables can be multi-dimensional and fix the dimension of each causal variable, which has also been explored in previous work \cite{DBLP:conf/cvpr/YangLCSHW21,iCITRIS}. This assumption enables the representation to learn more informative and specific latent codes to describe the factors of variation (i.e. x-position, y-position, etc.). We run experiments on datasets with continuous factors, but discrete flows \cite{discrete_flows} can be used for discrete factors. Our results suggest a component-wise correspondence between the learned and true causal factors.

\paragraph{Datasets.} We show the performance of our framework on three datasets, each consisting of four real-valued causal variables. The Pendulum dataset \cite{DBLP:conf/cvpr/YangLCSHW21} consists of causal variables with causal graph (pendulum angle $\to$ shadow length, shadow position) and (light position $\to$ shadow length, shadow position). The Flow dataset \cite{DBLP:conf/cvpr/YangLCSHW21} consists of variables with causal graph (ball size $\to$ water height), (hole position $\to$ water flow), and (water height $\to$ water flow). We also show experiments on a more complex 3D dataset of a robot arm interacting with colored buttons called CausalCircuit \cite{brehmer2022weakly}, which consists of variables with causal graph (robot arm $\to$ blue light intensity, green light intensity, and red light intensity), (blue light intensity $\to$ red light intensity), and (green light intensity $\to$ red light intensity).

\paragraph{Evaluation Metrics.} The DCI metric \cite{eastwood2018a} quantifies the degree to which ground-truth factors and learned latents are in one-to-one correspondence. We compute the DCI disentanglement ($D$) and completeness ($C$) scores, which are based on a feature importance matrix quantifying the degree to which each latent code is important for predicting each ground truth causal factor. The informativeness ($I$) score is the prediction error in the latent factors predicting the ground-truth generative factors and is constant ($I=0$) throughout all datasets and models, so we omit it for brevity. We train models with $3$ random seeds and select the median DCI score to report. We note that DCI is highly correlated with other disentanglement metrics, such as MCC, with strong connections to identifiability \cite{eastwood2023dcies}. To evaluate how changes in the generative factors affect the latent factors, we compute the interventional robustness score (IRS) \cite{pmlr-v97-suter19a}, which is similar to an $R^2$ value.

\paragraph{Implementation.} For the Pendulum ($6$K training and $1$K testing) and Flow ($6$K training and $2$K testing) datasets, we linearly increase the $\beta$ parameter throughout training from $0$ to $1$. We train for $9\cdot 10^3$ steps using a batch size of $64$. We use a Gaussian encoder and decoder with mean and variance computed by fully connected neural networks. For the CausalCircuit dataset ($35$K training and $10$K testing), we linearly increase $\beta$ from $0$ to $0.05$. We train for $3.5 \cdot 10^4$ steps using a batch size of $100$. We use a convolutional neural network architecture with $6$ layers and ReLU activation followed by a fully connected layer to estimate the mean and variance. The noise level for the variance of the Gaussian distribution of the conditional prior is controlled by $\sigma^2 \in \{0.01, 0.00001\}$. The structural causal flow and $\lam$ are implemented as affine form autoregressive flows with the slope and offset computed by fully connected 3-layer neural networks with $100$ unit hidden layers and ReLU activation. We set the learning rate to $0.001$ for all experiments. We set the dimension of each causal variable to $4$ for all datasets. Our code is available at \texttt{\url{https://github.com/Akomand/ICM-VAE}}.

\paragraph{Baselines.} We compare the performance of our approach, in terms of disentanglement and interventional robustness, with four baseline models: $\beta$-VAE \cite{higgins2017betavae} (unsupervised and acausal), iVAE \cite{pmlr-v108-khemakhem20a} (acausal), CausalVAE \cite{DBLP:conf/cvpr/YangLCSHW21} (causal), and SCM-VAE \cite{komanduri-scm-vae} (causal). The causal baselines propose relatively simplistic models that do not necessarily guarantee the disentanglement of causal factors.

\paragraph{Causal Disentanglement.} Our experiments show that learning diffeomorphic causal mechanisms, rather than linear SCM, and incorporating the causal structure to learn a bijective $\lam$ to estimate the parameters of the causally factorized distribution significantly improves the disentanglement and interventional robustness of learned causal factors compared with baselines, as shown in Table \ref{tab:full_supervision_disentanglement}. Consistent with our intuition, iVAE fails to disentangle the causal factors. The results indicate that ICM-VAE disentangles the causal factors and mechanisms almost perfectly. A high DCI disentanglement score indicates a permutation matrix mapping the latent factors to ground-truth generative factors in an ideal one-to-one mapping \cite{eastwood2023dcies}. Further, our model improves the interventional robustness of the representation, where interventions on ground-truth factors map to interventions on the corresponding learned factors.

\begin{table}
    \centering
    \footnotesize
    \setlength{\tabcolsep}{5.5pt}
    \begin{tabular}{lcccc}
        \toprule
        Dataset & Model & $D$ & $C$ & IRS \\
        \midrule
        Pendulum & $\beta$-VAE  & $0.182$ & $0.285$ & $0.449$ \\\
         & iVAE  & $0.483$ & $0.385$ & $0.670$ \\ & CausalVAE  & $0.885$ & $0.539$ & $0.817$ \\
         & SCM-VAE & $0.764$ & $0.475$ & $0.829$ \\ \rowcolor{ourmethod} & ICM-VAE (Ours) & $\mathbf{0.997}$ & $\mathbf{0.882}$ & $\mathbf{0.869}$ \\
        \midrule
        Flow & $\beta$-VAE  & $0.308$ & $0.332$ & $0.452$ \\
         & iVAE & $0.730$ & $0.481$ & $0.674$ \\ & CausalVAE & $0.819$ & $0.522$ & $0.707$ \\
         & SCM-VAE & $0.854$ & $0.483$ & $0.811$ \\ \rowcolor{ourmethod} & ICM-VAE (Ours)  & $\mathbf{0.988}$ & $\mathbf{0.598}$ & $\mathbf{0.893}$ \\
        \midrule
        CausalCircuit & $\beta$-VAE & $0.692$ & $0.442$ & $0.982$ \\
         & iVAE & $0.745$ & $0.541$ & $0.992$ \\ & CausalVAE & $0.886$ & $0.625$ & $0.994$ \\
         & SCM-VAE & $0.867$ & $0.652$ & $0.993$ \\ \rowcolor{ourmethod} & ICM-VAE (Ours) & $\mathbf{0.982}$ & $\mathbf{0.689}$ & $\mathbf{0.999}$ \\
        \bottomrule
    \end{tabular}
    \caption{Causal Disentanglement of ICM-VAE and baselines}
        \label{tab:full_supervision_disentanglement}
\end{table}

\paragraph{Counterfactual Generation.} We show counterfactual generated results of intervening on learned latent factors. Figure \ref{fig:cc_ctrfact} shows the CausalCircuit system and the result of intervening on the robot arm factor and propagating causal effects. We observe that the red light also turns on as the robot arm interacts with the blue or green lights. On the other hand, when the arm interacts with the red light, only the red light turns on and the other lights remain off. We observe a similar phenomenon in the Pendulum and Water Flow systems in Figure \ref{fig:pendflow_ctrfact}, which shows the result of intervening on causal factors and propagating effects. Intervening on the pendulum angle or light position has causal effects on the shadow. However, interventions on the shadow factors do not change the parent factors. For the counterfactual generation procedure, results from intervening on other causal factors, and iVAE latent traversals, refer to Appendix \ref{app:additional_exp}.

\section{Conclusion}
We propose ICM-VAE, a framework for causal representation learning under label supervision. We model causal mechanisms as flow-based transformations from noise to causal variables. We extend the idea of disentanglement to causal models and propose the notion of causal mechanism disentanglement. To this end, we design a causal disentanglement prior to causally factorize the distribution over causal variables. We theoretically show permutation-equivalent identifiability of the learned factors. Experimental results show that ICM-VAE almost perfectly disentangles the causal factors, improves interventional robustness, and generates consistent counterfactuals. Future work will incorporate causal discovery and disentanglement given partially observed labels.

\section*{Acknowledgements}

This work is supported in part by National Science Foundation under awards 1910284, 1946391 and 2147375, the National Institute of General Medical Sciences of National Institutes of Health under award P20GM139768, and the Arkansas Integrative Metabolic Research Center at University of Arkansas.

\bibliographystyle{named}
\bibliography{ijcai24}

\clearpage
\appendix
\makeatletter
\@addtoreset{theorem}{section}
\makeatother

\section*{Appendices}

\section{Theory}
\label{app:theory}

\subsection{Restatement and Proof of Theorem 1}\label{app:proof}

The following definitions and lemmas describe properties of exponential family sufficient statistics and the implication of causal mechanism permutation equivalence, which are used in our proof of Theorem 1.

\begin{definition}[Minimal Sufficient Statistic \normalfont{\cite{lachapelle2022disentanglement}}]
Given a parameterized distribution in the exponential family, we say its sufficient statistic $\T_i$ is minimal when there exists no $v \neq 0$ such that $v^T\T_i(z)$ is constant for all $z\in \mathcal{Z}$.
\end{definition}

\begin{definition}[Permutation-Scaling Matrix \normalfont{\cite{lachapelle2022disentanglement}}]
    A matrix is permutation-scaling if every row or column contains exactly one non-zero element.
\end{definition}

\begin{lemma}[\normalfont{\cite{lachapelle2022disentanglement}}]
    A sufficient statistic $\T: \mathcal{Z} \to \mathbb{R}^k$ is minimal if and only if there exist $z_{(0)}, \dots, z_{(k)}$ belonging to the support of $\mathcal{Z}$ such that the following $k$-dimensional vectors are linearly independent:
    \begin{equation}
        \T(z_{(1)}) - \T(z_{(0)}), \dots, \T(z_{(k)}) - \T(z_{(0)})
    \end{equation}
\end{lemma}

\begin{definition}
    \label{conditional_sufficient_statistic}
    A conditional sufficient statistic $\T(z | y): \mathcal{Z} \times \mathcal{Y} \to \mathbb{R}^k$ describes the sufficient statistics of the conditional distribution of $z$ induced as a result of conditioning on variable $y$. 
\end{definition}

\begin{definition}
\label{causal_perm_to_perm}
    For all $i, j\in \{1, \dots, n\}$, if $\T_i(z_i|z_{\textbf{pa}_i})$ and $\hat{\T}_i(z_j|z_{\textbf{pa}_j})$ are causal permutation-equivalent, then $z_{i}$ and $z_{j}$ are permutation-equivalent.
\end{definition}

We adapt the theory from \cite{lachapelle2022disentanglement} and \cite{pmlr-v108-khemakhem20a} and propose the following theorem for identifiability.

\begin{theorem}[\normalfont{Identifiability of ICM-VAE}] \label{thm1_app} Suppose that we observe data sampled from a generative model defined according to (8)-(12) with two sets of model parameters $\theta = (g, \T, \lam, G^z)$ and $\hat{\theta} = (\hat{g}, \hat{\T}, \hat{\lam}, \hat{G}^z)$. Suppose the following assumptions hold

\begin{enumerate}
    \item The set $\{x\in \mathcal{X} | \phi_{\xi}(x) = 0\}$ has measure zero, where $\phi_{\xi}$ is the characteristic function of the density $p_{\xi}$ defined in Eq. (9).
    \item The decoder $g$ is diffeomorphic onto its image.
    \item The sufficient statistics $\T_{i}$ are diffeomorphic.
    \item {\textbf{[Sufficient Variability]}} There exists $nk + 1$ distinct points $u_0, \dots, u_{nk}$ such that the matrix
    \begin{equation}
    \begin{split}
        L = (\lam(z_{\textbf{pa}_{(1)}}, u_{(1)}) - \lam(z_{(0)}, u_{(0)}), \dots, \\ \lam(z_{\textbf{pa}_{(nk)}}, u_{(nk)}) - \lam(z_{(0)}, u_{(0)}))
    \end{split}
    \end{equation}
    of size $nk \times nk$ is invertible, the ground-truth function $\lam$ is affected sufficiently strongly by each individual label $u_i$ and previously derived variables $z_{\textbf{pa}_i}$, and $\forall i$, $\lam_i(z_{\textbf{pa}_i}, u_i) \neq 0$.
\end{enumerate}
Then $\theta$ and $\hat{\theta}$ are causal mechanism permutation-equivalent, and the model $\hat{\theta}$ is causally disentangled. 
\end{theorem}

\textit{Proof Intuition.} 
The general idea of the proof is as follows. First, we use assumption 1 to show that the observed data distribution with noise is equivalent to the noiseless distribution. Then, using assumptions 2-4, we show the existence of an invertible linear transformation (in terms of the full-rank matrix $L$) between the learned and true causal factors, transforming both the sufficient statistic and mechanism $\lam$. Finally, by utilizing the causal graph (either fixed or learned), we show that the linear transformation must have a permutation and scaling structure and leads to causal mechanism equivalence due to the structure of the conditional sufficient statistics and causal mechanisms $\lam$. Thus, we show that the model is identifiable up to permutation and elementwise reparameterization. 

\proof To show that the set of parameters $\hat{\theta}$ is identifiable up to permutation, we break down the proof into five main steps. \\

\textbf{Step 1 (Equality of Denoised Distributions).} Firstly, we show that we can transform the equality of observed data distributions into a statement about the equality of noiseless distributions. Suppose we have two sets of parameters $\theta$ and $\hat{\theta}$ such that their marginal distributions are equivalent as follows:
\begin{equation}
    p_\theta (x | u) = p_{\hat{\theta}}(x|u)
\end{equation}
for all pairs $(x, u)$. Let $g^{-1} = s \circ a^{-1}$ be the encoding of causal factors $z$, where $a^{-1}$ is the encoding of the noise variables $\epsilon$. Then, we have the following

\begin{equation}
    \begin{split}
        \int_{\mathcal{Z}}& p_{\T, \lam}(z|u)p_{g}(x | z) \; dz \\ &= \int_{\mathcal{Z}} p_{\hat{\T}, \hat{\lam}}(z|u)p_{\hat{g}}(x|z) \; dz
    \end{split}
\end{equation}

\begin{equation}
    \begin{split}
        \int_{\mathcal{Z}} &p_{\T, \lam}(z|u)p_{\xi}(x - g(z)) \; dz \\ &= \int_{\mathcal{Z}} p_{\hat{\T}, \hat{\lam}}(z|u)p_{\xi}(x - \hat{g}(x)) \; dz 
    \end{split}
\end{equation}

\begin{equation}
    \begin{split}
        \int_{\mathcal{X}} & p_{\T, \lam}(g^{-1}(\bar{x})|u)\det J_{g^{-1}}(\bar{x}) p_{\xi}(x - \bar{x}) \; d\bar{x}  \\ &= \int_{\mathcal{X}} p_{\hat{\T}, \hat{\lam}}(\hat{g}^{-1}(\bar{x})|u)\det J_{\hat{g}^{-1}}(\bar{x})p_{\xi}(x - \bar{x}) \; d\bar{x} 
    \end{split}
\end{equation}

\begin{equation}
    \begin{split}
        \int_{\mathbb{R}^d} &p_{\T, \lam, g, u}(\bar{x}) p_{\xi}(x - \bar{x}) \; d\bar{x} \\ &= \int_{\mathbb{R}^d} p_{\hat{\T}, \hat{\lam}, \hat{g}, \hat{u}}(\bar{x})p_{\xi}(x - \bar{x}) \; d\bar{x} 
    \end{split}
\end{equation}

\begin{equation}
    (p_{\T, \lam, g, u} * p_{\xi})(x) = (p_{\hat{\T}, \hat{\lam}, \hat{g}, \hat{u}} * p_{\xi})(x)
\end{equation}

\begin{equation}
    \mathcal{F}[p_{\T, \lam, g, u}](w) \phi_{\xi}(w) = \mathcal{F}[p_{\hat{\T}, \hat{\lam}, \hat{g}, \hat{u}}](w) \phi_{\xi}(w) 
    \label{fourier_app}
\end{equation}

\begin{equation}
    \mathcal{F}[p_{\T, \lam, g, u}](w) = \mathcal{F}[p_{\hat{\T}, \hat{\lam}, \hat{g}, \hat{u}}](w)
    \label{post_conv}
\end{equation}

\begin{equation}
    p_{\T, \lam, g, u} = p_{\hat{\T}, \hat{\lam}, \hat{g}, \hat{u}}
    \label{marg}
\end{equation}

where $\mathcal{F}$ is the Fourier transform. Eq. (\ref{fourier_app}) and Eq. (\ref{marg}) use the fact that the Fourier transform is invertible and Eq. (\ref{post_conv}) is an application of the fact that the Fourier transform of a convolution is the product of their Fourier transforms. Thus, we have shown that if the distributions with added noise are the same, then the noise-free distributions must also be the same over all possible values $(x, u)$ within the support. \\

\textbf{Step 2 (Linear relationship).} Define $v = \hat{g}^{-1} \circ g: \mathcal{Z} \to \hat{\mathcal{Z}}$. By replacing Eq. (\ref{marg}) with the exponential form of the conditional prior from Eq. (11), we obtain the following:

\begin{align}
    p_{\T, \lam}(z | u) &= p_{\hat{\T}, \hat{\lam}}(z | u) \\
    p_{\T, \lam}(g^{-1}(x) | u)\det J_{g^{-1}}(x) &= p_{\hat{\T}, \hat{\lam}}(\hat{g}^{-1}(x) | u)\det J_{\hat{g}^{-1}}(x) 
\end{align}

\begin{equation}
    \begin{split}
         \prod_{i=1}^n h_i(g^{-1}_i(x)) \exp \Bigg[\sum_{j=1}^k T_{i, j}(g^{-1}_i(x) | g^{-1}_{\textbf{pa}_i}(x)) \lambda_{i, j}(z_{\textbf{pa}_i}, u_i) & \\- \psi_i(z_{\textbf{pa}_i}, u_i) \Bigg] \det J_{g^{-1}}(x)  =\\ \prod_{i=1}^n h_i(\hat{g}^{-1}_i(x)) \exp \Bigg[\sum_{j=1}^k \hat{T}_{i, j}(\hat{g}^{-1}_i(x) | g^{-1}_{\textbf{pa}_i}(x)) \hat{\lambda}_{i, j}(v(z_{\textbf{pa}_i}), u_i) \\ - \hat{\psi}_i(z_{\textbf{pa}_i}, u_i)\Bigg] \det J_{\hat{g}^{-1}}(x) \label{conditional_prior}
    \end{split}
\end{equation}

Taking the logarithm of both sides of Eq. (\ref{conditional_prior}) yields the following
\begin{equation}
\begin{split}
    &\log \det J_{g^{-1}}(x) + \sum_{i=1}^n \log  h_i(g^{-1}_i(x)) \\ &+ \sum_{j=1}^k T_{i, j}(g_i^{-1}(x)| g^{-1}_{\textbf{pa}_i}(x))\lambda_{i, j}(z_{\textbf{pa}_i}, u) - \psi_i(z_{\textbf{pa}_i}, u_i) \\
    &= \log \det J_{\hat{g}^{-1}}(x) + \sum_{i=1}^n \log \hat{h}_i(\hat{g}^{-1}_i(x)) \\ &+ \sum_{j=1}^k \hat{T}_{i, j}(\hat{g}_i^{-1}(x)| g^{-1}_{\textbf{pa}_i}(x))\hat{\lambda}_{i, j}(v(z_{\textbf{pa}_i}), u) -\hat{\psi}_i(z_{\textbf{pa}_i}, u_i)
\end{split}
\label{log_rel}
\end{equation}

Let $u_0, \dots, u_{nk}$ be the points provided by assumption 4 and define $\Delta\lam(z_{\textbf{pa}}, u) = \lam(z_{\textbf{pa}}, u) - \lam(z_0, u_0)$. The notation $z_{\textbf{pa}}$ indicates that each causal variable in $z$ is derived from its parents. We substitute each of the $u_{\ell}$ in the above equation to obtain $nk + 1$ distinct equations. Using $u_0$ as a pivot, we subtract the first equation for $u_0$ from the remaining $nk$ equations to obtain $\forall \ell\in 1, \dots, nk,$
\begin{equation}
\begin{split}
     &\T(g^{-1}(x))^T \Delta\lam(z_{\textbf{pa}_{\ell}}, u_{\ell}) \\&- \sum_i \psi_{i}(z_{\textbf{pa}_{\ell}},u_{\ell}) - \psi_{i}(z_0, u_0) \\= & \hat{\T}(\hat{g}^{-1}(x))^T \Delta\hat{\lam}(v(z_{\textbf{pa}_{\ell}}), u_{\ell}) \\&- \sum_i \hat{\psi}_{i}(z_{\textbf{pa}_{\ell}}, u_{\ell}) - \hat{\psi}_{i}(z_0, u_0) \label{vectorized_log_term}
\end{split}
\end{equation}

Now, let $L$ be the full-rank matrix described in the sufficient variability assumption (assumption 4), and $\hat{L}$ the matrix defined with respect to $\hat{\lam}$. Note that $\hat{L}$ is not guaranteed to be full-rank. Regrouping all normalizing constants $\psi$ into a term $b_{\ell} = d(z_{\textbf{pa}_{\ell}}, z_0, u_{\ell}, u_0)$ and letting $b$ be the vector of all $b_{\ell}$ for all $\ell\in \{1, \dots, nk\}$, we obtain the following:
\begin{equation}
    L^T\T(g^{-1}(x)) = \hat{L}^T \hat{\T}(\hat{g}^{-1}(x)) + b
\end{equation}
Since $L$ is assumed to be invertible, we can multiply by the inverse of $L^T$ on both sides to obtain the following
\begin{equation}
    \T(g^{-1}(x)) = A\hat{\T}(\hat{g}^{-1}(x)) + c
    \label{eq:linear_rel}
\end{equation}
where $A = (L^T)^{-1}\hat{L}$ and $c = (L^T)^{-1}b$. 

\textbf{Step 3 (Invertibility of $A$).} We show that $A$ is an invertible matrix. By Lemma 1, we have that the minimality of sufficient statistic $\T_i$ implies that the following set of vectors is linearly independent:
\begin{equation}
    \T_i(z_i^{(1)}) - \T_i(z_i^{(0)}), \dots, \T_i(z_i^{(k)}) - \T_i(z_i^{(0)})
\end{equation}
Define 
\begin{equation}
    z^{(0)} = [z_1^{(0)}, \dots, z_n^{(0)}]^T \in \mathbb{R}^n
\end{equation}
For all $i\in \{1, \dots, n\}$ and $p\in \{1, \dots, k\}$, define the vectors
\begin{equation}
    z^{(p, i)} = [z_1^{(0)}, \dots, z_{i-1}^{(0)}, z_i^{(p)}, z_{i+1}^{(0)} \dots, z_n^{(0)}]^T \in \mathbb{R}^n
\end{equation}
Now, for $1 \leq p \leq k$ and $i\in \{1, \dots, n\}$, we consider the following difference
\begin{equation}
    \T(z^{(p, i)}) - \T(z^{(0)}) = A[\hat{\T}(z^{(p, i)}) - \hat{\T}(z^{(0)})]
    \label{ele_diff}
\end{equation}
where the LHS is a vector filled with zeros except for the block corresponding to $\T_i(z_i^{(p, i)}) - \T(z_i^{(0)})$. Define
\begin{equation}
    \Delta \T^{(i)} = [\T(z^{(1, i)}) - \T(z^{(0)}) \dots \T(z^{(k, i)}) - \T(z^{(0)})]
\end{equation}
and
\begin{equation}
    \Delta \hat{\T}^{(i)} = [\hat{\T}(z^{(1, i)}) - \hat{\T}(z^{(0)}) \dots \hat{\T}(z^{(k, i)}) - \hat{\T}(z^{(0)})]
\end{equation}
Then, we have that the columns of both these are linearly independent and all rows are filled with zeros except the block of rows $\{(i-1)k + 1, \dots, ik\}$. So, writing Eq. (\ref{ele_diff}) in matrix form and grouping all components, we have the following
\begin{equation}
    [\Delta \T^{(1)}, \dots, \Delta \T^{(n)}] = A [\Delta \hat{\T}^{(1)}, \dots, \Delta \hat{\T}^{(n)}]
\end{equation}
Thus, we have a block diagonal matrix of size $nk\times nk$. Since each block is invertible, $[\Delta \T^{(1)}, \dots, \Delta \T^{(n)}]$ must be invertible. This implies that $A$ must be invertible. Thus, Eq. (\ref{eq:linear_rel}) and the invertibility of $A$ imply that $\theta \sim_A \hat{\theta}$.

\begin{figure*}
\centering
    \includegraphics[scale=0.35]{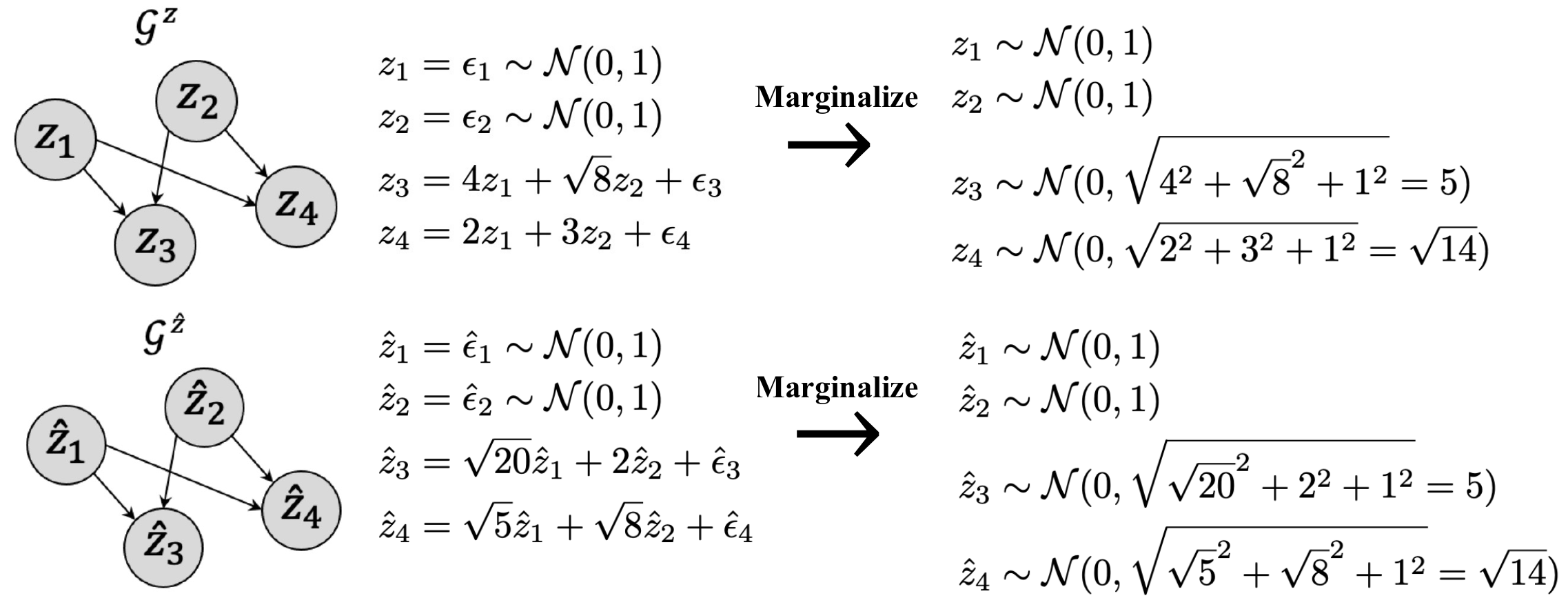}
    \caption{Counterexample to traditional disentanglement}
    \label{fig:counterex}
\end{figure*}

\textbf{Step 4 (Linear relationship of natural parameters).} In addition to showing the linear relationship between the sufficient statistic, we also show the linear relationship linking $\lam$ and $\hat{\lam}$. Define $v = \hat{g}^{-1} \circ g: \mathcal{Z} \to \hat{\mathcal{Z}}$. That is, there exists a diffeomorphism between the learned and ground-truth factors. We rewrite Eq. (\ref{log_rel}) as follows:
\begin{equation}
\begin{split}
    &\T(g^{-1}(x))^T \lam(z_{\textbf{pa}}, u) \\ &= \hat{\T}(\hat{g}^{-1}(x))^T \hat{\lam}(v(z_{\textbf{pa}}), u) + \kappa(z_{\textbf{pa}}, u) + \gamma(z)
\end{split}
\end{equation}
where we combine all terms depending only on $z$ into $\gamma$ and those depending only on $u$ into $\kappa$. Now, we can rewrite the above as follows using the linear relationship between sufficient statistics:
\begin{equation}
\begin{split}
    &\hat{\T}(\hat{g}^{-1}(x))^T A^T \lam(z_{\textbf{pa}}, u) + c^T\lam(z_{\textbf{pa}}, u) \\ &= \hat{\T}(\hat{g}^{-1}(x))^T \hat{\lam}(v(z_{\textbf{pa}}), u) + \kappa(z_{\textbf{pa}}, u) + \gamma(z)
\end{split}
\end{equation}
\begin{equation}
\begin{split}
    &\hat{\T}(\hat{g}^{-1}(x))^T A^T \lam(z_{\textbf{pa}}, u) \\ &= \hat{\T}(\hat{g}^{-1}(x))^T \hat{\lam}(v(z_{\textbf{pa}}), u) + \bar{\kappa}(z_{\textbf{pa}}, u) + \gamma(z)
\end{split}
\end{equation}
where $\bar{\kappa}$ absorbs all $u$-dependent terms. We can simplify this equality to the following:
\begin{equation}
\begin{split}
    &\hat{\T}(\hat{g}^{-1}(x))^T (A^T \lam(z_{\textbf{pa}}, u) - \hat{\lam}(v(z_{\textbf{pa}}), u)) \\ &= \bar{\kappa}(z_{\textbf{pa}}, u) + \gamma(z)
\end{split}
\end{equation}
\begin{equation}
\begin{split}
    &\T(\hat{g}^{-1}(x))^T (A^T \lam(z_{\textbf{pa}}, u) - \hat{\lam}(v(z_{\textbf{pa}}), u)) \\ &= \bar{\kappa}(z_{\textbf{pa}}, u) + \gamma(\hat{g}^{-1}(x))
\end{split}
\end{equation}
Taking the finite difference between distinct values $z$ and $\bar{z}$ yields
\begin{equation}
\begin{split}
    &[\T(z) - \T(\bar{z})]^T (A^T\lam(z_{\textbf{pa}}, u) - \hat{\lam}(v(z_{\textbf{pa}}), u)) \\ &= \gamma(\hat{g}^{-1}(x)) - \gamma(\hat{g}^{-1}(\bar{x}))
\end{split}
\end{equation}
Now, we can construct an invertible matrix $[\Delta \T^{(1)} \dots \Delta \T^{(n)}]$ such that 
\begin{equation}
\begin{split}
    [\Delta \T^{(1)} \dots \Delta \T^{(n)}]^T &(A^T\lam(z_{\textbf{pa}}, u) - \hat{\lam}(v(z_{\textbf{pa}}), u)) \\ &= [\Delta\gamma^{(1)} \dots \Delta\gamma^{(n)}]
\end{split}
\end{equation}
Due to this invertibility, we can simplify the above to obtain
\begin{equation}
    A^T\lam(z_{\textbf{pa}}, u)  + \gamma = \hat{\lam}(v(z_{\textbf{pa}}), u) 
\end{equation}
where 
\begin{equation}
    \gamma  = - [\Delta \T^{(1)} \dots \Delta \T^{(n)}]^{-T}[\Delta\gamma^{(1)} \dots \Delta\gamma^{(n)}]
\end{equation}
We can rewrite this as follows to yield the equivalence of $\lam$ and $\hat{\lam}$
\begin{equation}
    A^T\lam(z_{\textbf{pa}}, u) + \gamma = \hat{\lam}(v(z_{\textbf{pa}}), u)
\end{equation}

\textbf{Step 5 (Permutation Equivalence).} To show that $A$ is a permutation-scaling matrix, we have to show that any two columns cannot have nonzero entries on the same row. 

\begin{itemize}
    \item If $G^z$ is fixed, we are done since the trivial permutation always holds. Since the decoder is assumed to be diffeomorphic, there is a point-wise nonlinearity between each corresponding factor of the representation. Thus, a bijective mapping establishes a component-wise reparameterization (scaling) and trivial permutation.

\item If $G^z$ is learned, then for a learned sparse graph $\hat
{G}^z$ if the following holds
\begin{equation}
    \pi(\hat{G}^z) = G^z
\end{equation}
then we still achieve permutation equivalence by permuting the causal graph.
\end{itemize}

We conclude that $A$ must be a causal permutation-scaling matrix. Since $\lam$ captures the causal dependencies between factors of $z$, we have that the following must be true
\begin{equation}
    \T_i(z_i | z_{\textbf{pa}_i}) = A_{ij} \hat{\T}_j(z_j | z_{\textbf{pa}_j})
\end{equation}
Thus, $\T$, $\hat{\T}$ and $\lam$, $\hat{\lam}$ must be causal mechanism permutation-equivalent, respectively, and $z$ is causally disentangled. Therefore, $z$ is disentangled, and we have that $\theta \sim_P \hat{\theta}$, where $P=A$ is a permutation-scaling matrix.

\subsection{Traditional disentanglement cannot guarantee independent causal mechanisms equivalence} \label{app:counterex}
In this section, we provide a counterexample to show that the traditional notion of disentanglement cannot capture the equivalence of causal mechanisms. For example, consider the running example of the Pendulum system. We have four causal factors that are causally related. Let $z$ denote the true factors of variation and $\hat{z}$ denote the learned factors, where each $z_i$ and $\hat{z}_i$ correspond to the same causal variable. For the sake of simplicity, consider the example in Figure \ref{fig:counterex}.

Observe that the causal mechanisms learned are different than the true causal mechanisms. We have the following equivalent marginal distribution for true and learned factors:

\begin{equation*}
\begin{split}
    p(z) &= p(z_1)p(z_2)p(z_3|z_1, z_2)p(z_4|z_1, z_2) \\ &\approx p(\hat{z}_1)p(\hat{z}_2)p(\hat{z}_3|\hat{z}_1, \hat{z}_2)p(\hat{z}_4|\hat{z}_1, \hat{z}_2) = p(\hat{z})
\end{split}
\end{equation*}

However, traditional disentanglement does not imply equivalence of all individual causal mechanisms of true and learned factors. In the above example, the true SCM consists of different mechanisms than the learned SCM, but both yield the same marginal distribution. This example violates the causal mechanism permutation equivalence and causal disentanglement but satisfies traditional disentanglement. We claim that learning a model that achieves equivalence of causal mechanisms from the perspective of the ICM principle better captures disentanglement in the causal setting.

\begin{figure*}[t!]
    \centering
    \begin{subfigure}[t]{0.3\textwidth}
        \centering
        \includegraphics[scale=0.65]{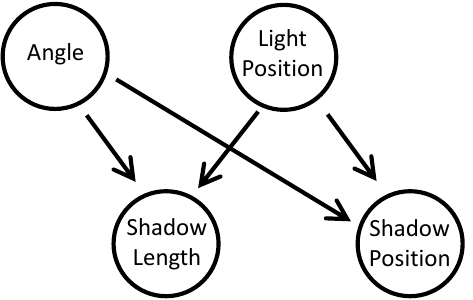}
        \caption{Pendulum}
        \label{fig:pendu}
    \end{subfigure}
    ~ 
    \begin{subfigure}[t]{0.3\textwidth}
        \centering
        \includegraphics[scale=0.65]{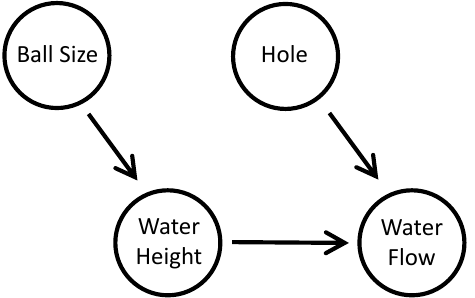}
        \caption{Water Flow}
        \label{fig:wf}
    \end{subfigure}
        ~ 
    \begin{subfigure}[t]{0.3\textwidth}
        \centering
        \includegraphics[scale=0.65]{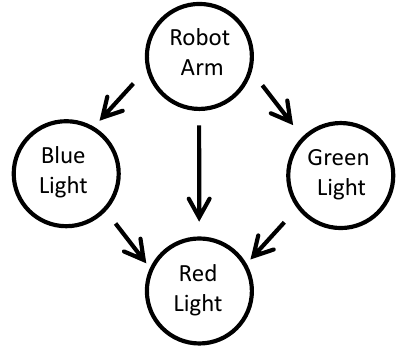}
        \caption{CausalCircuit}
        \label{fig:cc}
    \end{subfigure}\caption{Causal Graphs of Datasets}
    \label{causal_graphs}
\end{figure*}

\subsection{Derivation of ICM-VAE ELBO} \label{app:elbo_der}
We aim to push the variational posterior distribution $q_{\phi}(\epsilon, z | x, u)$ to the true joint posterior distribution $p_{\theta}(\epsilon, z | x, u)$. Formally, the goal is to minimize the KL divergence as follows:

$\mathcal{D}(q_{\phi}(\epsilon, z | x, u), p_{\theta}(\epsilon, z | x, u))$
\begin{align}
     &= \int\int q_{\phi}(\epsilon, z | x, u) \log \frac{q_{\phi}(\epsilon, z | x, u)}{p_{\theta}(\epsilon, z | x, u)} \; d\epsilon\; dz \\
     &= \int\int q_{\phi}(\epsilon, z | x, u) \log \frac{q_{\phi}(\epsilon, z| x, u) p_{\theta}(x, u)}{p_{\theta}(\epsilon, z, u, x)} \; d\epsilon\; dz \\
     \begin{split}
     &= \int\int q_{\phi}(\epsilon, z | x, u) \Big[ \log p_{\theta}(x, u) \\& \qquad\qquad+ \log \frac{q_{\phi}(\epsilon, z | x, u)}{p_{\theta}(\epsilon, z, u, x)} \Big] \; d\epsilon\; dz 
     \end{split}\\
     \begin{split}
     &= \int\int q_{\phi}(\epsilon, z | x, u) \log p_{\theta}(x, u) \\ & \qquad\qquad+ q_{\phi}(\epsilon, z | x, u) \log \frac{q_{\phi}(\epsilon, z | x, u)}{p_{\theta}(\epsilon, z, u, x)}  \; d\epsilon\; dz 
     \end{split} \\
     \begin{split}
     &= \log p_{\theta}(x, u) \\ & \qquad+ \int\int q_{\phi}(\epsilon, z | x, u) \log \frac{q_{\phi}(\epsilon, z | x, u)}{p_{\theta}(\epsilon, z, u, x)}  \; d\epsilon\; dz 
     \end{split} \\
     \begin{split}
     &= \log p_{\theta}(x, u) \\ &+ \int\int q_{\phi}(\epsilon, z | x, u) \log \frac{q_{\phi}(\epsilon, z | x, u)}{p_{\theta}(x | \epsilon, z, u)p_{\theta}(\epsilon, z, u)}  \; d\epsilon\; dz 
     \end{split} \\
     \begin{split}
     &= \log p_{\theta}(x, u) + \mathbb{E}_{\epsilon, z \sim q_{\phi}(\epsilon, z | x, u)} \Big[\log \frac{q_{\phi}(\epsilon, z | x, u)}{p_{\theta}(\epsilon, z, u)} \\ 
     &\qquad\qquad- \log p_{\theta}(x | \epsilon, z, u) \Big] 
     \end{split}\\
     \begin{split}
     &= \log p_{\theta}(x, u) + \mathcal{D}(q_{\phi}(\epsilon, z | x, u) || p_{\theta}(\epsilon, z | u)) \\ & \qquad\qquad- \mathbb{E}_{\epsilon, z \sim q_{\phi}(\epsilon, z | x, u)} \Big[\log p_{\theta}(x | \epsilon, z, u) \Big]
     \end{split}
\end{align}

\noindent
So, we have the following:
\begin{equation}
\begin{split}
    \mathcal{D}(q_{\phi}(\epsilon, z | x, u) || &p_{\theta}(\epsilon, z | x, u)) = \log p_{\theta}(x, u) \\&+ \mathcal{D}(q_{\phi}(\epsilon, z | x, u) || p_{\theta}(\epsilon, z | u)) \\ &- \mathbb{E}_{\epsilon, z \sim q_{\phi}(\epsilon, z | x, u)} \Big[\log p_{\theta}(x | \epsilon, z, u) \Big]
\end{split}
\end{equation}

\noindent
Rearranging, we can simply the objective to the following:

\begin{equation}
\begin{split}
    &\log p_{\theta}(x, u) - \mathcal{D}(q_{\phi}(\epsilon, z | x, u) || p_{\theta}(\epsilon, z | x, u)) \\ &=  \mathbb{E}_{\epsilon, z \sim q_{\phi}(\epsilon, z | x, u)} \Big[\log p_{\theta}(x | \epsilon, z, u) \Big] \\ &- \mathcal{D}(q_{\phi}(\epsilon, z | x, u) || p_{\theta}(\epsilon, z | u))
\end{split}
\end{equation}

\noindent
This implies that
\begin{equation}
\begin{split}
    \log p_{\theta}(x, u) &- \mathcal{D}(q_{\phi}(\epsilon, z | x, u) || p_{\theta}(\epsilon, z | x, u)) \\ &\leq \log p_{\theta}(x, u)
\end{split}
\end{equation}

\noindent
Putting everything together yields the following evidence lower bound (ELBO)
\begin{equation}
\begin{split}
    \underbrace{\log p_{\theta}(x, u)}_{\text{Evidence}} &\geq \overbrace{\mathbb{E}_{\epsilon, z \sim q_{\phi}(\epsilon, z | x, u)} \Big[\log p_{\theta}(x | \epsilon, z, u) \Big]}^{\text{Likelihood}} \\ &- \underbrace{\mathcal{D}(q_{\phi}(\epsilon, z | x, u) || p_{\theta}(\epsilon, z, u))}^{\text{KL Term}}
\end{split}
\end{equation}

\noindent
Now, since $\epsilon$ and $z$ are related by a diffeomorphism, we can simplify the objective as follows.
\begin{equation}
\begin{split}
    \log p_{\theta}(x, u) \geq &\mathbb{E}_{\epsilon, z \sim q_{\phi}(\epsilon, z | x, u)} \Big[\log p_{\theta}(x | \epsilon, z, u) \Big] \\ &- \mathcal{D}(q_{\phi}(\epsilon | x, u) || p_{\theta}(\epsilon)) \\&-\mathcal{D}(q_{\phi}(z | x, u) || p_{\theta}(z|u))
\end{split}
\end{equation}

\noindent
obtained by the following derivation

$\log p_{\theta}(x, u) \geq$
\begin{align}
\begin{split}
    &\mathbb{E}_{\epsilon, z \sim q_{\phi}(\epsilon, z | x, u)} \Big[\log p_{\theta}(x | \epsilon, z, u) \Big] \\&- \mathbb{E}_{\epsilon, z \sim q_{\phi}(\epsilon, z | x, u)} \Big[\log \frac{q_{\phi}(\epsilon, z | x, u)}{p_{\theta}(\epsilon, z)}\Big] 
    \end{split}\\
    &= \mathbb{E}_{\epsilon, z \sim q_{\phi}(\epsilon, z | x, u)} \Big[\log p_{\theta}(x | \epsilon, z, u)  -  \log \frac{q_{\phi}(\epsilon, z | x, u)}{p_{\theta}(\epsilon, z)} \Big] \\
    \begin{split}
    &= \mathbb{E}_{\epsilon, z \sim q_{\phi}(\epsilon, z | x, u)} \Big[\log p_{\theta}(x | \epsilon, z)  -  \log q_{\phi}(\epsilon, z | x, u) \\ &\qquad+ \log p_{\theta}(\epsilon, z) \Big] 
    \end{split}\\
    \begin{split}
    &= \mathbb{E}_{\epsilon, z \sim q_{\phi}(\epsilon, z | x, u)} \Big[\log p_{\theta}(x | \epsilon) + \log p_{\theta}(x| z) \\ & \qquad-  \log q_{\phi}(\epsilon, z | x, u) + \log p_{\theta}(\epsilon) + \log p_{\theta}(z | u) \Big] 
    \end{split}
    \\
    \begin{split}
    &= \mathbb{E}_{\epsilon, z \sim q_{\phi}(\epsilon, z | x, u)} \Big[\log p_{\theta}(x | \epsilon) + \log p_{\theta}(x| z)  \\ & \qquad-  \log q_{\phi}(\epsilon | x, u) - \log q_{\phi}(z | x, u) \\& \qquad+ \log p_{\theta}(\epsilon) + \log p_{\theta}(z | u) \Big]
    \end{split}
\end{align}

\section{Additional Experimental Details}
\label{app:exp}

\subsection{Dataset Details}
\label{app:dataset_det}

\underline{\textbf{Pendulum}}. The Pendulum dataset \cite{DBLP:conf/cvpr/YangLCSHW21} is a synthetic dataset that consists of $7$K images with resolution $96\times 96\times 4$ generated by $4$ ground-truth causal variables: $u_1=$ pendulum angle, $u_2=$ light position, $u_3=$ shadow length, and $u_4=$ shadow position, which are continuous values. Each causal variable is determined from the following process with nonlinear functions. The causal graph is shown in Figure \ref{fig:pendu}.

\begin{equation*}
    u_1 \sim U(-45, 45);\; \qquad \theta = u_1 * \frac{\pi}{200}
\end{equation*}

\begin{equation*}
    u_2 \sim U(60, 145);\; \qquad \phi = u_2 * \frac{\pi}{200}
\end{equation*}

\begin{equation*}
    x = 10 + 9.5\sin\theta
\end{equation*}

\begin{equation*}
    y = 10 - 9.5\cos\theta
\end{equation*}

\begin{align*}
    u_3 &= \max(3, \Big|\frac{-0.5 - (10.5-10\tan\phi)}{\tan\phi} \\ &- \frac{-0.5 - (y-x\tan\phi)}{\tan\phi}\Big|) \\ &= \max(3, \Big|\frac{(-10.5+y) + (10-x)\tan\phi}{\tan\phi}\Big|) \\ &= \max(3, \Big|9.5\frac{\cos\theta}{\tan\phi} + 9.5\sin\theta\Big|)
\end{align*}

\begin{align*}
    u_4 &=  \frac{1}{2}\Big(\frac{-0.5 - (10.5-10\tan\phi)}{\tan\phi} \\ &+\frac{-0.5 - (y-x\tan\phi)}{\tan\phi}\Big) \\ &= \frac{1}{2}\Big(\frac{(-11.5-y) + (10+x)\tan\phi}{\tan\phi}\Big) \\ &= \frac{-11 + 4.75\cos\theta}{\tan\phi} + (10 + 4.75\sin\theta)
\end{align*}

\noindent
\underline{\textbf{Flow}}. The Flow dataset \cite{DBLP:conf/cvpr/YangLCSHW21} is a synthetic dataset that consists of $8$K images with resolution $96\times 96\times 4$ generated by $4$ ground-truth causal variables: $u_1=$ ball radius, $u_2=$ water height, $u_3=$ hole position, and $u_4=$ water flow, which are continuous values. The causal graph is shown in Figure \ref{fig:wf}.
\\

\noindent
\underline{\textbf{Causal Circuit}}. The Causal Circuit dataset is a new dataset created by \cite{brehmer2022weakly} to explore research in causal representation learning. The dataset consists of $512\times 512\times 3$ resolution images generated by $4$ ground-truth latent causal variables: robot arm position, red light intensity, green light intensity, and blue light intensity. The images show a robot arm interacting with a system of buttons and lights. The data is rendered using an open-source physics engine. The original dataset consists of pairs of images before and after an intervention has taken place. For this work, we only utilize observational data of either the before or after system. The data is generated according to the following process:
\begin{align*}
    v_R &= 0.2 + 0.6 * \text{clip}(u_2 + u_3 + b_R, 0, 1) \\
    v_G &= 0.2 + 0.6 * b_G \\
    v_B &= 0.2 + 0.6 * b_B \\
    u_4 &\sim \text{Beta}(5v_R, 5 * (1 - v_R)) \\
    u_3 &\sim \text{Beta}(5v_G, 5 * (1 - v_G)) \\
    u_2 &\sim \text{Beta}(5v_B, 5 * (1 - v_B)) \\
    u_1 &\sim U(0, 1)
\end{align*}
where $b_R$, $b_G$, and $b_B$ are the pressed state of buttons that depends on how far the button is touched from the center, $u_1$ is the robot arm position, and $u_2$, $u_3$, and $u_4$ are the intensities of the blue, green, and red lights, respectively. The causal graph is shown in Figure \ref{fig:cc}. From this generative process, we selectively choose only images for which the causal graph is satisfied (the robot arm's position and the downstream effects). For example, the robot arm appearing over the green button, green button lit up, and red button lit up is consistent with the assumption that the robot arm position causes changes in which buttons light up according to the causal graph. The filtered dataset consists of roughly $35$K training samples and $10$K testing samples.

\subsection{Counterfactual Generation}
\label{app:counterfactual_generation}
Following \cite{Pearl09}, the process for obtaining counterfactual predictions consists of three steps

\begin{enumerate}
    \item \textbf{Abduction:} given an observation $x$, we infer the distribution over the latent variable $\epsilon$ via $\epsilon = a^{-1}(x)$. In the context of ICM-VAE, we have that $z = g^{-1}(x) = f^{RF}(\epsilon)$.
    \item \textbf{Action:} substitute the values of $z$ with values based on the counterfactual query $z_{z_j \leftarrow \alpha}$
    \item \textbf{Prediction:} using the modified model and the value of $z$, compute (decode to) the value of $x$, the consequence of the counterfactual.
\end{enumerate}

We clarify that, unlike the abduction step from traditional counterfactual inference, in our counterfactual generation procedure, we first derive noise terms from the observed high dimensional data and derive the causal factors from the noise terms. This effectively implies we obtain the original noise terms associated with the latent causal factors.

\subsection{Additional Experimental Results}
\label{app:additional_exp}
\begin{figure*}
    \centering
    \includegraphics[width=\textwidth]{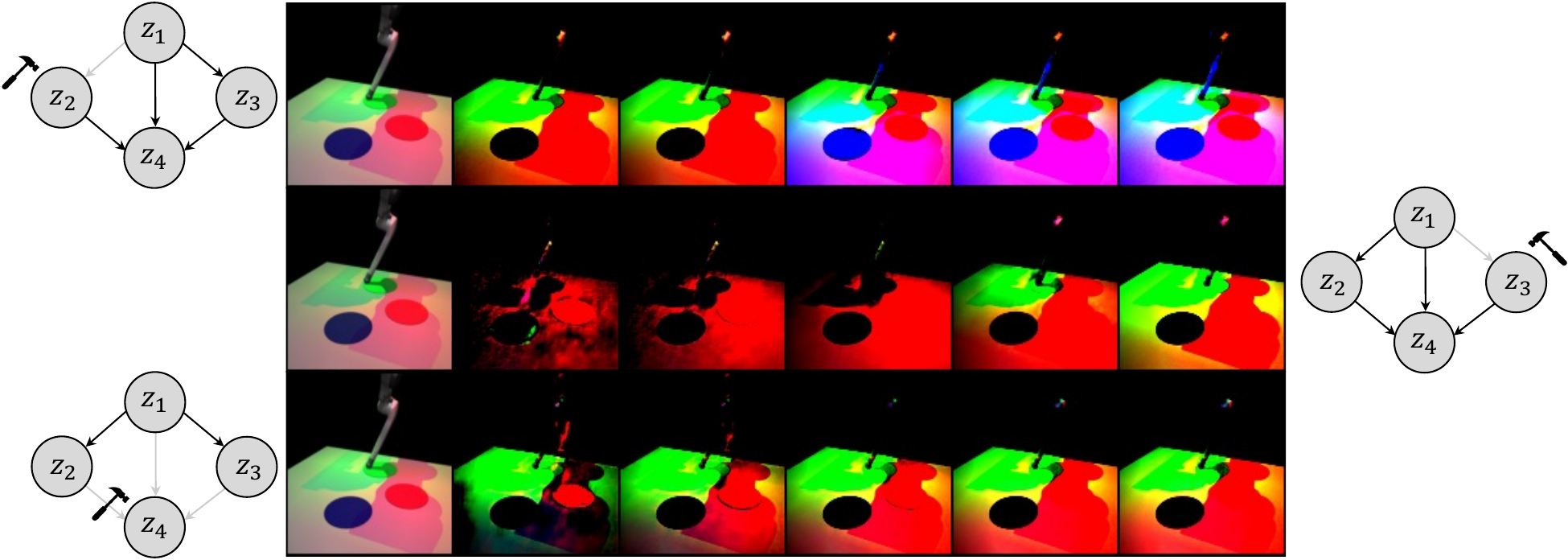}
    \caption{ICM-VAE counterfactuals generated after intervening on blue light (top), green light (middle), and red light (bottom), respectively.}
    \label{fig:icm_vae_counter_ex_appendix}
\end{figure*}

\begin{figure*}
    \centering
    \includegraphics[scale=0.5]{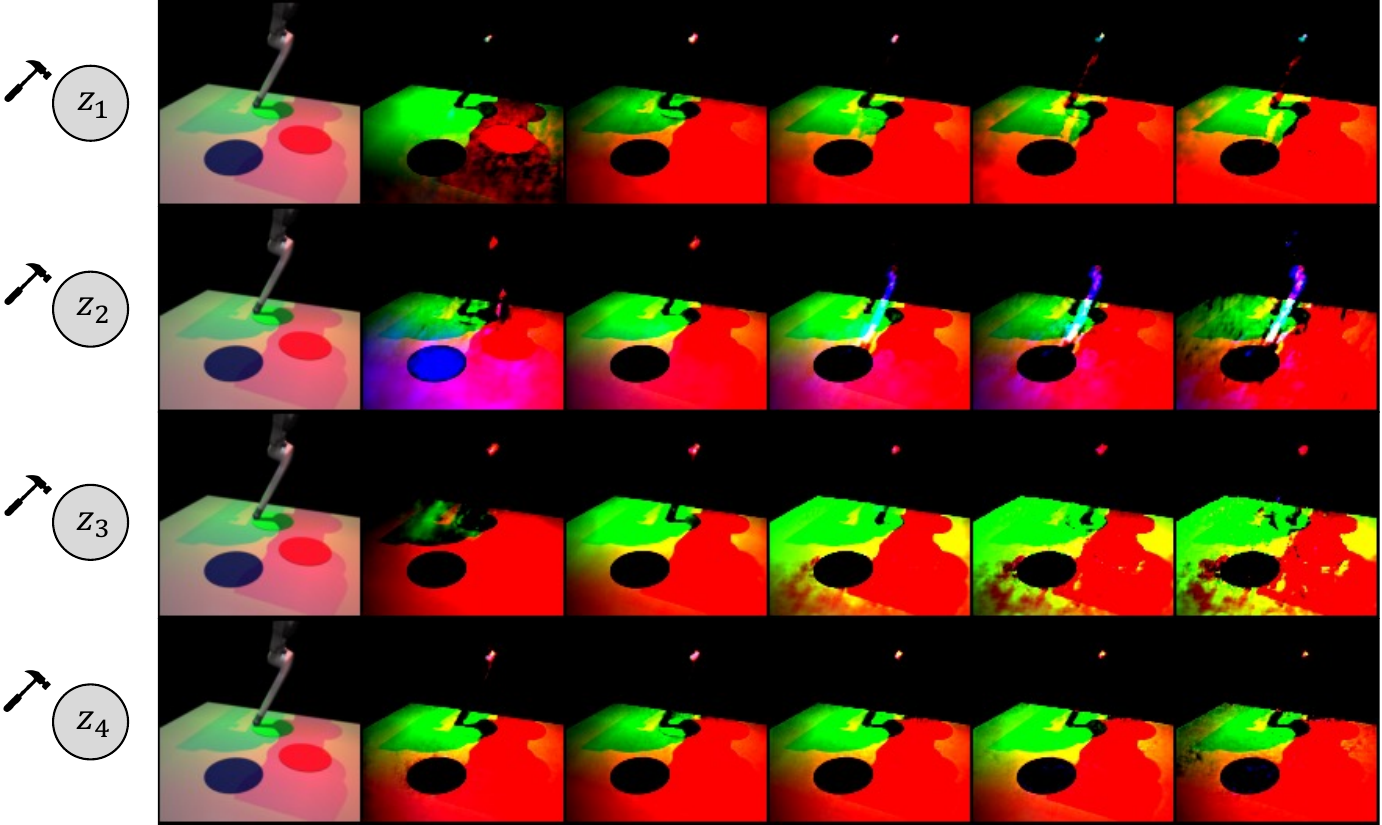}
    \caption{iVAE generated images after latent traversal on each causal variable: robot arm (first), blue light (second), green light (third), and red light (fourth), respectively.}
    \label{fig:ivae_counter_ex_appendix}
\end{figure*}

\begin{figure*}
    \centering
    \includegraphics[scale=0.6]{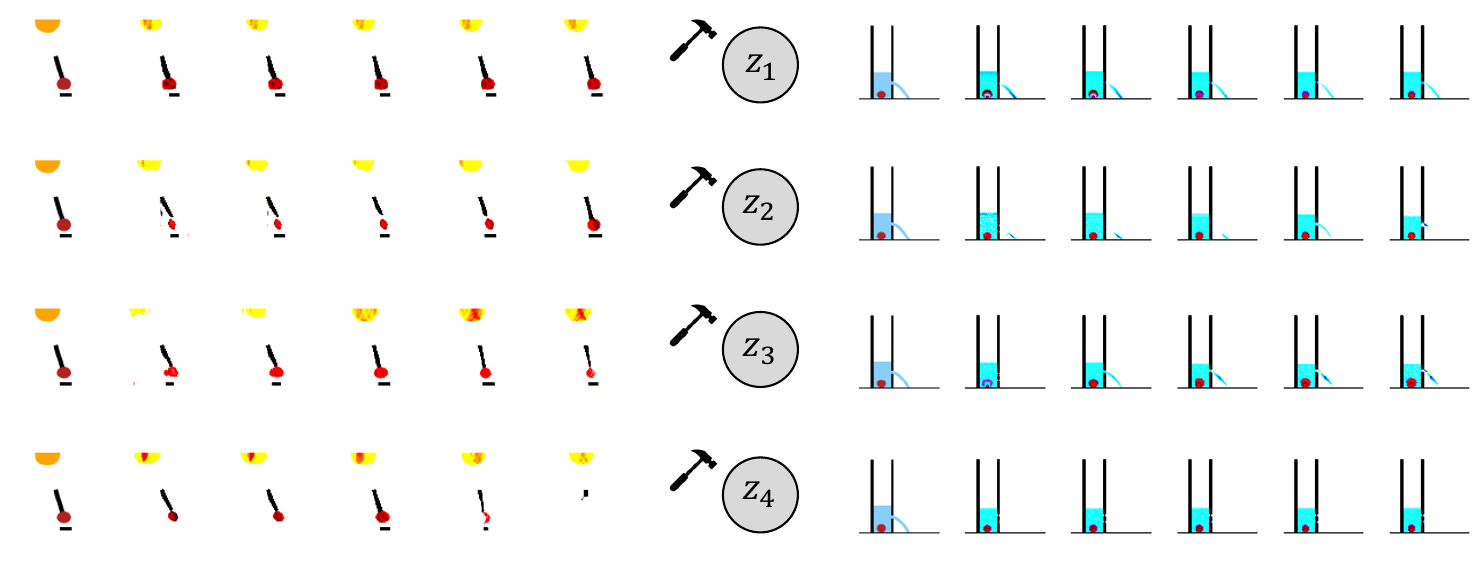}
    \caption{iVAE generated images after latent traversal on each causal variable for Pendulum and Water Flow datasets, respectively.}
    \label{fig:ivae_counter_ex_appendix_pendflow}
\end{figure*}

We include additional counterfactual images generated using ICM-VAE on other causal variables from the CausalCircuit dataset in Figure \ref{fig:icm_vae_counter_ex_appendix}. Observe that we can manipulate $z_2$ in an isolated fashion to change the brightness of the blue light and the brightness of the red light after computing causal effects. When intervening on $z_3$, we see a change in the green light and a change in the brightness of the red light. Finally, after intervening on $z_4$, we see a change in the brightness of the red light, but not its causal parents, which is consistent with the intervention. Further, we include iVAE-generated images after latent traversals on causal variables for the CausalCircuit dataset (Figure \ref{fig:ivae_counter_ex_appendix}) and the Pendulum/Water Flow datasets (Figure \ref{fig:ivae_counter_ex_appendix_pendflow}). We can see that iVAE is not able to disentangle the causal factors and not capable of generating counterfactual images according to the assumed causal model since iVAE is an acausal method.

\subsection{Extension to Discrete-valued Factors}
In this work, we focus on formulating ideas in causal disentanglement and representation learning and use synthetic datasets with continuous-valued variables for evaluation. However, an extension of our framework to datasets with discrete-valued variables (such as CelebA \cite{7410782}) would take the form of parameterizing causal mechanisms using discrete flows \cite{discrete_flows}. 

\subsection{Baselines}
We compare ICM-VAE with four baselines: two acausal and two causal. 

\textbf{$\beta$-VAE} \cite{higgins2017betavae} is an unsupervised disentanglement method that aims to promote disentanglement in the latent space by encouraging the latent representation to be more factorized. However, $\beta$-VAE is unable to effectively disentangle the factors of variation to a high degree, which is consistent with the claim from \cite{pmlr-v97-locatello19a} that unsupervised disentanglement is not possible without additional inductive biases. 

\textbf{iVAE} \cite{pmlr-v108-khemakhem20a} unifies nonlinear ICA and the VAE to develop a framework for learning identifiable representations using auxiliary information in the form of a conditionally factorial prior. However, the framework of iVAE assumes independent factors of variation, which is often an impractical assumption. Due to this assumption, iVAE is unable to disentangle causally related factors. 

\textbf{CausalVAE} \cite{DBLP:conf/cvpr/YangLCSHW21} and \textbf{SCM-VAE} \cite{komanduri-scm-vae} extended the iVAE framework for causally related factors of variation. CausalVAE utilizes a prior that still assumes mutual independence of the factors of variation. Further, CausalVAE assumes a simple linear SCM, which is unrealistic in practice. SCM-VAE builds on this work and consists of a post-nonlinear additive noise SCM and a label-specific causal prior. However, the causal prior proposed still does not induce a causal factorization of latent factors. Thus, CausalVAE and SCM-VAE are also unable to properly disentangle the causal factors. 

\subsection{Compute}
 We run our experiments on an Ubuntu 20.04 workstation with eight NVIDIA Tesla V100-SXM2 GPUs with 32GB RAM.

\section{Discussion on Causal Discovery}
\label{app:discovery_discussion}
Similar to \cite{shen_disentangled_2021,liang2023causal}, in this work, we assume the latent causal structure is known to theoretically formulate the idea of causal mechanism disentanglement. In principle, our theory works just the same for a discovered causal graph as long as it is recovered up to some permutation of the true graph. This can be formalized as a graph isomorphism between the learned and ground-truth causal graphs \cite{brehmer2022weakly}. Incorporating causal discovery methods, such as NOTEARS \cite{NOTEARS} or GraN-DAG \cite{Lachapelle2020Gradient-Based}, could be integrated heuristically in the form of adding a penalty to terms in the VAE loss objective to enforce sparsity and acyclicity as follows
\begin{equation}
    \mathcal{L}_{total} = \mathcal{L}_{ICM-VAE} + H(A) + \|A\|_0
\end{equation}
where $H(A) = tr[(I + \alpha A \odot A)]^n - n = 0$ is the acyclicity constraint and $\|\cdot\|_0$ enforces the sparsity of the DAG. We can alternatively use the $\|\cdot\|_1$ for sparsity to ensure a differentiable objective. Similar to \cite{NOTEARS}, we can utilize the augmented Lagrangian to optimize the joint loss objective. However, since we use flow-based models to parameterize causal mechanisms, the topological ordering of the causal graph is important. Since our focus in this work is not causal discovery, we assume knowledge of the causal graph from domain knowledge. We assume a flexible diffeomorphic general nonlinear structural causal model. Most causal discovery methods require restricting assumptions, such as linear (or post-nonlinear) additive noise models, to be practically applied. In this work, we motivate the need for more sophisticated causal discovery methods for generalized SCMs that may not necessarily assume some parametric form. A promising candidate is Jacobian-based causal discovery \cite{reizinger2022multivariable}, which leverages the structure of nonlinear ICA methods and the Jacobian of causal mechanisms to facilitate causal discovery. Additionally, other causal discovery algorithms could be used heuristically with a variety of different assumptions \cite{vowels2021d,causaldisc_survey}. We look to explore this direction in future work.

\end{document}